\documentclass{article}

\usepackage{microtype}
\usepackage{graphicx}
\usepackage{subcaption}
\usepackage{booktabs} 
\usepackage{xurl}

\usepackage{hyperref}

\usepackage[accepted]{icml2026}

\usepackage{amsmath}
\usepackage{amssymb}
\usepackage{mathtools}
\usepackage{amsthm}
\usepackage{enumitem}

\usepackage[capitalize,noabbrev]{cleveref}

\theoremstyle{plain}

\theoremstyle{definition}

\theoremstyle{remark}

\usepackage[textsize=tiny]{todonotes}

\icmltitlerunning{Collaborative Threshold Watermarking}

\begin{document}

\twocolumn[
  \icmltitle{Collaborative Threshold Watermarking}



  \icmlsetsymbol{equal}{*}

  \begin{icmlauthorlist}
    \icmlauthor{Tameem Bakr}{equal,MBZUAI}
    \icmlauthor{Anish Ambreth}{equal,MBZUAI}
    \icmlauthor{Nils Lukas}{MBZUAI}
  \end{icmlauthorlist}

  \icmlaffiliation{MBZUAI}{Department of Machine Learning, MBZUAI, Abu Dhabi, UAE}

  \icmlcorrespondingauthor{Tameem Bakr}{tameem.bakr@mbzuai.ac.ae}

  \icmlkeywords{Machine Learning, ICML}

  \vskip 0.3in
]



\printAffiliationsAndNotice{}  

\begin{abstract}
In federated learning (FL), $K$ clients jointly train a model without sharing raw data. 
Because each participant invests data and compute, clients need mechanisms to later prove the provenance of a jointly trained model. 
Model watermarking embeds a hidden signal in the weights, but naive approaches either do not scale with many clients as per-client watermarks dilute as $K$ grows, or give any individual client the ability to verify and potentially remove the watermark. 
We introduce $(t,K)$-threshold watermarking: clients collaboratively embed a shared watermark during training, while only coalitions of at least $t$ clients can reconstruct the watermark key and verify a suspect model. 
We secret-share the watermark key $\tau$ so that coalitions of fewer than $t$ clients cannot reconstruct it, and verification can be performed without revealing $\tau$ in the clear.
We instantiate our protocol in the white-box setting and evaluate it on image
classification tasks on both IID and non-IID partitions, as well as  language models fine-tuning
setting.
Our watermark remains detectable at scale ($K=128$) with minimal accuracy loss and stays above the detection threshold ($z\ge 4$) under attacks including adaptive fine-tuning using up to 20\% of the training data. Code is available at \url{https://github.com/tameemalaa/collaborative-threshold-watermark}.
\end{abstract}

\section{Introduction}

Federated learning (FL) enables multiple parties to train machine learning models collaboratively without sharing raw data. 
FL has already been deployed at scale, for example, in Google Keyboard \citep{48270, yang2018appliedfederatedlearningimproving}, Apple's voice recognition \citep{granqvist20_interspeech}, and other privacy-sensitive applications \citep{NVIDIAFederatedLearningHealthcare, 10.1609/aaai.v37i13.26847, Muzellec2024.12.06.627138}.
Recent work by \citet{sani2024futurelargelanguagemodel,rui2024openfedllm} envisions training large language models (LLMs) from scratch via FL to democratize model training and ownership.
\begin{minipage}[t]{\columnwidth}
  \centering
  \includegraphics[width=\linewidth]{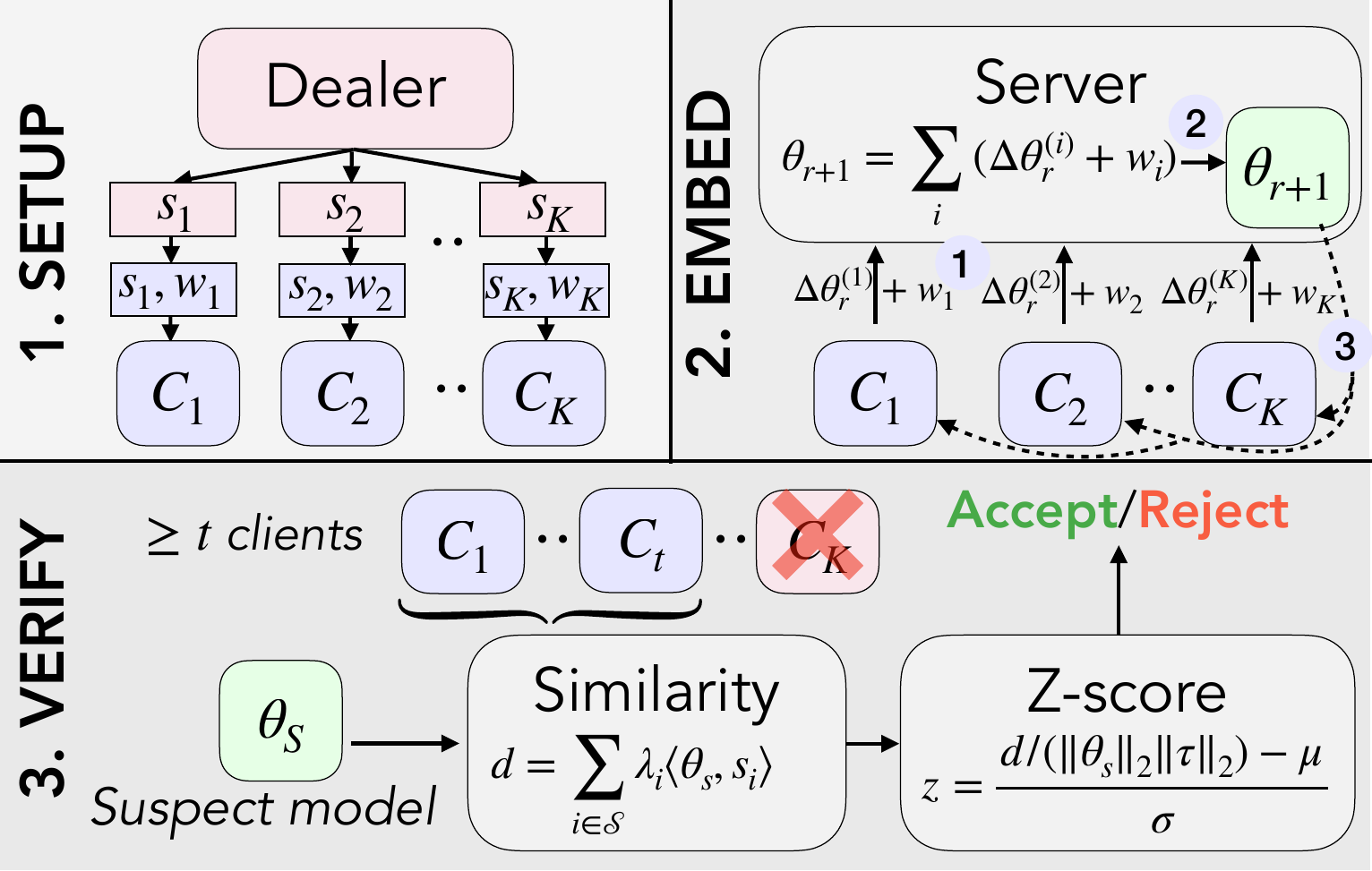}
  \captionof{figure}{
  An overview of collaborative threshold watermarking. $\textsc{Setup}$ is a one-time procedure to distribute Shamir shares $s_i$ to all clients from which they derive additive shares $w_i$. $\textsc{Embed}$ modifies the FL algorithm to embed our watermark, $\textsc{Verify}$ allows any coalition of $\geq t$ clients to compute the watermark test statistic.}
  \label{fig:firstpage-rightcol}
\end{minipage}

Training large models consumes vast compute and data resources.
However, once a model is jointly trained in FL, any client could redistribute it without the consent of any other client. 
Clients must trust each other \emph{not} to leak the model before engaging in the FL protocol. 
However, FL often operates in settings with limited trust (e.g., without contractual obligations), so \emph{trustless}, distributed mechanisms are needed to verify a model's provenance.

Model watermarking \citep{10.1145/3078971.3078974,10.5555/3277203.3277324} is a solution to verify a model's provenance.
Watermarking embeds a hidden signal that can later be detected by accessing the model using a secret watermarking key. 
Watermarking methods for FL must (i) preserve the model's utility, (ii) be reliably detectable, and (iii) be robust against unauthorized removal attempts~\citep{10.3389/fdata.2021.729663,lukas2021sokrobustimageclassification}.
Naive approaches fail at scale: if each client embeds its own watermark, the signal each client can embed into the model diminishes as $K$ grows. 
If all clients own the watermark key, then any individual client can verify and potentially remove the watermark, undermining robustness.
A protocol is needed where $K$ clients embed collaboratively, but verification requires a coalition of at least $t\leq K$ clients.

Threshold verification is useful in settings where ownership is collective: for example, in a healthcare FL consortium of five hospitals, choosing $t=3$ prevents any single hospital from unilaterally claiming provenance or removing the watermark, while still enabling majority-based verification.

We propose the first method for \emph{$(t,K)$-threshold model watermarking}. 
Our protocol combines (i) secret sharing, so that any coalition of at least $t$ clients can reconstruct a watermark key $\tau$ for verification, and (ii) secure aggregation, so that clients can embed shares $w_k$ that sum to $\tau$ during training with untrusted servers.
Secure aggregation hides individual client updates, but the server still observes the aggregated global model trajectory. 
Our goal is to prevent any party from reconstructing $\tau$ unless they hold at least $t$ shares.
Verification is performed in the white-box setting, computing the test statistic directly from shares without reconstructing the secret key $\tau$ in the clear, and computing a calibrated, one-sided $z$-score on a suspect model.
Across CIFAR-10, CIFAR-100 \citep{krizhevsky2009learning}, and Tiny ImageNet \citep{Le2015TinyIV}, we show that our approach scales to many clients (we test up to $K=128$), when a baseline watermark diminishes as the number of clients grows and falls below the detection threshold once $K \ge 16$. Additional experiments on CIFAR-100 under balanced Dirichlet
label skew shows that watermark detectability remains stable under non-IID
statistical heterogeneity, and a GPT-2 Small \citep{radford2019language} experiment on WikiText-2 \citep{merity2017pointer}
indicates that the method also extends beyond image classification. Our
watermark has a small impact on model utility and is robust to a wide range 
of both adaptive and non-adaptive removal attacks when access to training
data is limited.
Our watermark has a small impact on model accuracy and is robust to a wide range of both adaptive and non-adaptive removal attacks when access to training data is limited.

\subsection{Contributions.}
Our contributions can be summarized as follows: 
 \begin{itemize}[itemsep=0pt, topsep=0pt]
    \item We propose the first $(t,K)$-threshold watermark for FL that scales to many clients and ensures only $\geq t$ clients can collectively verify the presence of a watermark.
     \item We empirically show that our watermark has a negligible impact on the model's accuracy and is reliably detectable via a one-sided, calibrated $z$-test.

    \item Our watermark remains detectable under pruning up to $90\%$, 4-bit quantization, and adaptive fine-tuning with up to 20\% of the training data. 
 \end{itemize}

\section{Background}
We describe federated learning (FL), secure aggregation and model watermarking.
Then we review cryptographic primitives such as commitments and secret sharing.
\subsection{Federated Learning (FL)}
FL enables multiple clients to collaboratively train a shared model without exchanging raw data~\citep{9464278}.
Each client $i$ has a dataset of size $n_i$, and the global model $\theta$ is trained to minimize the overall weighted loss
\begin{align}
F(\theta) = \frac{1}{n} \sum_{i=1}^K n_i \, F_i(\theta),
\end{align}
where $F_i(\theta)$ is the loss on client $i$'s local data and $n = \sum_{i=1}^K n_i$. 
The server maintains a global model $\theta_r$ at round $r$.
Training proceeds across many rounds: (1) The server sends the current global model to clients, (2) clients update it on their local data, and (3) the server aggregates these updates and sends the updated model back to the clients. 

\paragraph{Secure Aggregation}
\label{sec:secagg}
Secure aggregation is a key primitive in  FL for protecting client privacy. While FL avoids sharing raw data, individual client updates can still leak sensitive information if observed by the server. Secure aggregation mitigates this risk by ensuring that the server learns only an aggregate over client updates, not any individual contribution.
We write $y \leftarrow \textsc{SecAgg}(\{x_k\}_{k=1}^K)$ for a secure aggregation protocol that outputs $y=\sum_{k=1}^K x_k$ to the server, while keeping each individual $x_k$ hidden from the server.
We treat this as a standard FL building block and refer to known constructions such as that of \citet{bonawitz2016practical}. Such protocols incur communication overhead linear in the number of clients, modest additional local computation, and are robust to client dropouts up to a fixed threshold. Since our protocol invokes secure aggregation twice only over model-sized vectors, compared to once in normal FL training, its communication and computation costs are asymptotically equivalent to those of FL training.

\subsection{Model Watermarking}
\label{sec:bg-watermark}
A model watermark~\citep{10.1145/3078971.3078974} is a hidden signal that can be extracted from the model using a secret watermarking key. 
Formally, any watermarking method is defined by the following three algorithms:
\begin{itemize}[itemsep=0pt, topsep=0pt]
    \item $\tau \leftarrow \mathrm{SETUP}()$: Samples a secret key and prepares the information needed for embedding and verification. 
    \item $\theta_w \leftarrow \mathrm{EMBED}(\theta, \tau)$: Given a model's parameters $\theta$ and a secret key $\tau$, return watermarked parameters.
    \item $ y \leftarrow \mathrm{VERIFY}(\theta_s, \tau)$: Detect the watermark in a \emph{suspect} model $\theta_s$ using the key $\tau$, and output a z-score which rejects the null hypothesis that the detected signal is present due to random chance. 
\end{itemize}
A \emph{white-box} watermark requires that \textsc{Verify} has access to the model's parameters for verification, whereas a black-box watermark requires only API access.
By definition, black-box schemes are also verifiable in the white-box setting. 

\subsection{Commitment Schemes}
A \textit{commitment scheme}~\citep{brassard1988minimum,katz2020moderncrypto} 
is a two-phase protocol between a sender and receiver. 
It consists of a 
\textsc{Commit} algorithm to fix a value $m$ with randomness $r$, producing a 
commitment $\mathcal{C}$, and an \textsc{Open} algorithm where the sender reveals $(m,r)$ 
for verification. It has two characteristics: (i) \textbf{Hiding:} Commitments reveal nothing about $m$. (ii) \textbf{Binding:} It is infeasible to open the same commitment to two different values.
We write $\mathcal{C}\leftarrow \textsc{Commit}(\tau;\rho)$ to denote that we publish a public nonce $\rho$ and a commitment $\mathcal{C}$ to the watermark key $\tau$.

\subsection{Secret Sharing}
A \emph{$(t,n)$-threshold} secret sharing scheme~\citep{stinson2005,katz2020moderncrypto} splits a secret \(s\) into shares $s_1,\dots,s_n$ such that any $|\mathcal{S}|\ge t$ shares reconstruct $s$, while the joint distribution of any $|\mathcal{S}|<t$ shares is independent of $s$.

\paragraph{Shamir’s threshold scheme.}
In Shamir’s classical $(t,n)$-threshold scheme \citep{shamir1979secret}, the dealer samples a random degree-$(t-1)$ polynomial $P(x)$ over a finite field $\mathbb{F}_q$ with $P(0)=s$, where $s$ is the secret.
Each participant $i$ is assigned a distinct nonzero $x_i \in \mathbb{F}_q$ and receives the share $(x_i, s_i)$ where $s_i = P(x_i)$.
Any set of $t$ shares uniquely determines $P(x)$, while any set of fewer than $t$ shares reveals no information about $s$.
We denote share generation by $\{s_i\}_{i=1}^n \leftarrow \textsc{ShamirShare}(s,t,n)$ (with fixed public evaluation points, e.g., $x_i=i$), and reconstruction by $s \leftarrow \textsc{ShamirReconstruct}(\{s_i\}_{i\in\mathcal{S}})$ for any $|\mathcal{S}|\ge t$.

\paragraph{Additive (embedding) shares.}
We call vectors $w_1,\dots,w_K$ \emph{additive shares} of a secret $\tau$ if $\sum_{i=1}^K w_i=\tau$.
We highlight that if all clients hold Shamir shares of a secret $\tau$, they can cooperate to derive embedding shares via public Lagrange coefficients, while retaining Shamir’s $(t,n)$-threshold secrecy for verification.
Specifically, for fixed public evaluation points $\{x_i\}_{i=1}^K$, the Lagrange coefficient
\begin{align}
\label{eq:lagrange}
\lambda_i &= \prod_{j\neq i} \frac{0 - x_j}{x_i - x_j},
\qquad \text{such that } \tau = \sum_{i=1}^K \lambda_i s_i .
\end{align}
is a publicly computable constant that depends only on the evaluation points.
Using~\eqref{eq:lagrange}, each client locally computes its embedding share as $w_i=\lambda_i s_i$.

\section{Threat Model}

We consider a standard FL setup: 
$K$ clients want to train a joint model without revealing their raw data, and each client contributes (i) data and (ii) computation. 
Clients want the ability to collectively verify the presence of a watermark, for example, to retain ownership or to deter unauthorized use of the model.

\paragraph{Trusted Dealer.} 
We consider both settings where (i) a trusted dealer is available (which we focus on in the main paper), and (ii) where a trusted dealer is unavailable.
In the latter case, clients need to invoke a dealer-free distributed key generation algorithm described in the Appendix.  

\paragraph{Clients and Server.}
Each client holds private local datasets, and the client set is fixed across rounds.
All clients are \emph{honest-but-curious}, meaning they follow the protocol during training, but may form coalitions of $<t$ clients to remove the watermark post-training.
We assume that clients do not trust the server, e.g., since a client could act as the server.
The goal is that the server does not learn the secret shares and that $\tau$ cannot be reconstructed without at least $t$ client shares.
Each client's objective is to support collaborative verification and resist watermark removal by any coalition of fewer than $t$ clients while maintaining utility. 

\paragraph{Adversary.}
We consider a white-box adversary with access to the entire training trajectory, i.e., all intermediate model checkpoints.
The adversary knows the secret shares of any coalition of $<t$ clients, and they have auxiliary data not used during FL, which they can use for further fine-tuning. 
Unless stated otherwise, we assume the attacker’s auxiliary data is labeled and drawn from the same distribution as the FL training data.
We focus on post-training watermark-removal attacks, and malicious training-time adversaries are out of scope for our work.
The adversary's objective is to obtain a model (i) with high accuracy\footnote{Our adversary lacks sufficient data to train a high-utility model from scratch. Otherwise, there is no need to engage in FL.} (ii) that does not contain the watermark.
\paragraph{Design Goals.} Our watermarking method should (i) allow shared ownership, where only a coalition of $\geq t$ clients can verify the watermark in a model.
We want the watermark to (ii) scale with an arbitrary number of clients $K$, and (iii) operate in the presence of an untrusted server that must not learn the secret watermark key $\tau$.
(iv) The watermark must be robust against removal attacks after FL training, and (v) have a low false positive rate on models trained without knowledge of our watermarking method. 

\section{Conceptual Approach}
The key idea is to use Shamir secret sharing so that only coalitions of size $\ge t$ can reconstruct $\tau$, while individual clients can still embed \emph{shares} of $\tau$ during training under secure aggregation. 
We instantiate this idea as three components, mirroring the watermarking primitives of Section~\ref{sec:bg-watermark}: a one-time \textsc{Setup} that distributes Shamir shares of $\tau$ to the $K$ clients, an \textsc{Embed} procedure that integrates these shares into the FedAvg update under secure aggregation, and a reconstruction-free \textsc{Verify} step that evaluates the watermark test directly from shares without ever materializing $\tau$ in the clear.

\subsection{Watermark Setup} 

\paragraph{Trusted Dealer.} 
\Cref{alg:setup_dealer} implements our \textsc{Setup} procedure that runs once at the beginning of the FL process among all $K$ clients.
The goal is to distribute verification and embedding shares of the secret key $\tau$ to all clients. 
We sample the watermark key as $\tau\sim \mathcal{N}(0,I_d)$. 

Because the aggregation server is untrusted with $\tau$, it never receives $\tau$ in the clear. 
Instead, after the setup, each client $k$ holds a Shamir share $s_k$ of $\tau$ and derives an \emph{embedding share} $w_k$ using the (public) Lagrange coefficient $\lambda_k$ for evaluation at $0$ over the full client set:
\begin{align}
w_k \;\leftarrow\; \lambda_k \cdot s_k \quad\text{so that}\quad \sum_{k=1}^K w_k = \tau.
\label{eq:embed_share}
\end{align}
For fixed public evaluation points (e.g., $x_k=k$), the coefficients $\{\lambda_k\}_{k=1}^K$ are fixed constants and can be precomputed once.
This lets clients embed shares locally under \textsc{SecAgg} (Section~\ref{sec:secagg}) while the server learns only the aggregated watermark contribution.

\begin{algorithm}[t]
    \caption{Trusted dealer setup (One-time operation).}
    \label{alg:setup_dealer}
    \begin{algorithmic}[1]
        \REQUIRE $K$ clients, threshold $t$, public evaluation points $\{x_k\}_{k=1}^K$
        \STATE Dealer samples watermark key $\tau$ and public nonce $\rho$
        \STATE Publish $(\rho, \mathcal{C} \leftarrow \textsc{Commit}(\tau; \rho))$
	        \STATE $\{s_k\}_{k=1}^K \leftarrow \textsc{ShamirShare}(\tau,t,K)$ \hfill (client shares)
	        \FOR{each client $k$}
	            \STATE Send $s_k$ to $k$ over an authenticated, private channel
	            \STATE $\lambda_k \leftarrow \prod_{\substack{j=1\\ j\neq k}}^K \frac{0-x_j}{x_k-x_j}$ \COMMENT{Lagrange Coefficient}
	            \STATE $w_k \leftarrow \lambda_k s_k$ (so $\sum_{k=1}^K w_k=\tau$)
	        \ENDFOR
	        \STATE Dealer deletes $\tau$
    \end{algorithmic}
\end{algorithm}
\paragraph{Dealer-free setup.}
If a trusted dealer is unavailable, clients can run a dealer-free distributed key generation (DKG) protocol to obtain Shamir shares of an implicit random $\tau$, revealing it to any party (Appendix~\ref{app:setup_dist}).
At a high level, each client $k$ samples a degree-$(t-1)$ polynomial $P_k$ with random constant term and privately sends evaluations $P_k(x_i)$ to each client $i$, who sums the received values to obtain its share. 
This requires $O(K^2)$ authenticated point-to-point messages (each carrying one fixed-point vector share in $\mathbb{Z}_q^d$) and $O(K^2 d)$ total communication.\footnote{Appendix~\ref{app:setup_dist:overhead} reports empirical per-client runtime and communication/computation overhead for the dealer-free DKG.}

\subsection{Watermark Embedding}
At each round $r$, client $k$ performs local training to obtain
$\theta^{(k)}_r$ and computes the local model update
\[
\Delta \theta^{(k)}_r := \theta^{(k)}_r - \theta_{r-1}.
\]
A naive embedding strategy would add a fixed-strength watermark
perturbation in every round. However,
client update magnitudes can vary substantially across rounds
and clients due to data heterogeneity, optimizer dynamics, and
learning-rate schedules. Fixed-strength embedding can therefore
be unstable, either degrading model utility or becoming
undetectable when update norms fluctuate.

To stabilize the embedding strength, we adaptively scale the
watermark magnitude based on recent update norms. Each client
maintains an exponential moving average (EMA) of its update
magnitudes. Let $\text{ema}^{(r)}_k$ denote the EMA tracker for
client $k$ at round $r$, updated with decay parameter $\beta$ as
\begin{equation}
\text{ema}^{(r)}_k
= \beta \cdot \text{ema}^{(r-1)}_k
+ (1 - \beta)\cdot \lVert \Delta \theta^{(k)}_r \rVert_2 .
\label{eq:ema}
\end{equation}
Each client then calibrates the strength of its watermark
perturbation as
\begin{equation}
\text{scale}_k
= c \cdot \lVert \Delta \theta^{(k)}_r \rVert_2
\cdot \text{ema}^{(r)}_k ,
\label{eq:scale}
\end{equation}
where $c$ is a global watermark strength hyperparameter.

To preserve the additive-share structure of the watermark and
prevent the server from learning individual scaling factors,
clients submit $\text{scale}_k$ via secure aggregation. The
server obtains only the global scaling factor
\begin{equation}
\text{scale}_{\text{total}}
\leftarrow
\textsc{SecAgg}\big(\{\text{scale}_k\}_{k=1}^K\big)
=
\sum_{k=1}^K \text{scale}_k ,
\label{eq:scale-secagg}
\end{equation}
which is broadcast to all clients. Using a single global
$\text{scale}_{\text{total}}$ is necessary to ensure that the sum of scaled shares holds to a scaled $\tau$; per-client scaling would break this additive structure. This procedure adds one secure aggregation
step per round, incurring modest communication overhead.

\paragraph{Per-client embedding.} Clients keep $\tau$ secret-shared and derive embedding shares $w_k$ from their Shamir shares $s_k$ as in Eq.~\ref{eq:embed_share}. After local training, clients contribute $\text{scale}_k$ to \textsc{SecAgg} (Section~\ref{sec:secagg}) so the server learns only $\text{scale}_{\text{total}}=\sum_k \text{scale}_k$, which it broadcasts. 
Each client forms a watermarked model $u_k \leftarrow \theta_r^{(k)} + \text{scale}_{\text{total}}\,w_k$ and contributes $u_k$ to \textsc{SecAgg}. The server learns only the aggregate and sets:
\begin{align}
\theta_{r} &= \frac{1}{K}\,\textsc{SecAgg}\Big(\big\{u_k\big\}_{k=1}^K\Big)
\label{eq:trustless}
\end{align}
Because $\sum_k w_k=\tau$, the magnitude of the watermark drift per round is $(\text{scale}_{\text{total}}/K)\,\tau$.

\paragraph{Client subsampling and dropouts.}
If only a subset of the clients $S_r\subseteq\{1,\dots,K\}$ participates in round $r$, the same idea applies as long as $|S_r|\ge t$, which is that each participating client $k\in S_r$ computes the round-specific Lagrange coefficient $\lambda_k^{(S_r)}$ for evaluation at $0$ over points $\{x_i\}_{i\in S_r}$ and sets $w_k^{(S_r)} \leftarrow \lambda_k^{(S_r)} s_k$.
Then $\sum_{k\in S_r} w_k^{(S_r)}=\tau$, and the aggregate update under \textsc{SecAgg} yields the same watermark direction (with $K$ replaced by $|S_r|$ in the averaging). Under full participation, $\lambda_k^{(S_r)}=\lambda_k$ is fixed and can be precomputed once.
If $|S_r|<t$, clients skip watermark embedding for that round. 

\paragraph{Weighted FedAvg.}
If the server applies weighted aggregation, i.e., 
\begin{equation}
\theta_r = \sum_{k \in S_r} a_{k,r} \bigl(\theta_r^{(k)} + \text{scale}_{\mathrm{total},r} w_k\bigr),
\label{eq:weighted-fedavg}
\end{equation}
with public weights \(a_{k,r} > 0\) such that \(\sum_{k \in S_r} a_{k,r} = 1\), each participating client \(k \in S_r\) rescales its embedding share to \(\tilde{w}_{k,r} \leftarrow \frac{1}{a_{k,r}} w_k\). Then, as shown in Eq.~\eqref{eq:weighted-fedavg}, we obtain
\begin{equation}
\sum_{k \in S_r} a_{k,r} \tilde{w}_{k,r}
=
\sum_{k \in S_r} a_{k,r} \frac{1}{a_{k,r}} w_k
=
\sum_{k \in S_r} w_k.
\label{eq:rescaled-sum}
\end{equation}

So the aggregated update preserves the same watermark direction as in the unweighted case (cf. Eq.~\eqref{eq:rescaled-sum}). As before, if \(|S_r| < t\), clients skip watermark embedding for that round.

%
%
%
%

\subsection{White-box Watermark Verification}
\label{sec:verification}
The simplest procedure to verify the watermark reconstructs the key $\tau$ from Shamir shares $\{ s_i \}_{i \in \mathcal{S}}$ via polynomial interpolation, but then the secret is leaked in the clear. 
However, reconstructing the key is not required since our test depends only on the inner product $\langle \theta_s,\tau\rangle$.
A coalition can compute it directly from shares without ever materializing $\tau$.
For a coalition $\mathcal{S}$ with $|\mathcal{S}|\ge t$ and Lagrange coefficients $\{\lambda_i\}_{i\in\mathcal{S}}$ for evaluation at $0$, we have $\tau = \sum_{i\in\mathcal{S}} \lambda_i s_i$, hence
\begin{align}
    \langle \theta_s,\tau\rangle = \sum_{i\in\mathcal{S}} \lambda_i \langle \theta_s, s_i\rangle.
\end{align}
Each client $i$ locally computes the scalar $\langle \theta_s, s_i\rangle$ and the coalition sums the weighted scalars via secure aggregation to obtain $\langle \theta_s,\tau\rangle$. 
Together with $\|\theta_s\|_2$ and a fixed $\|\tau\|_2$ implied by the key distribution, this suffices to compute $\cos(\theta_s,\tau)$ and the final $z$-score. 

\begin{figure*}[t]
\centering
\begin{subfigure}{0.32\textwidth}
  \centering
  \includegraphics[width=\linewidth]{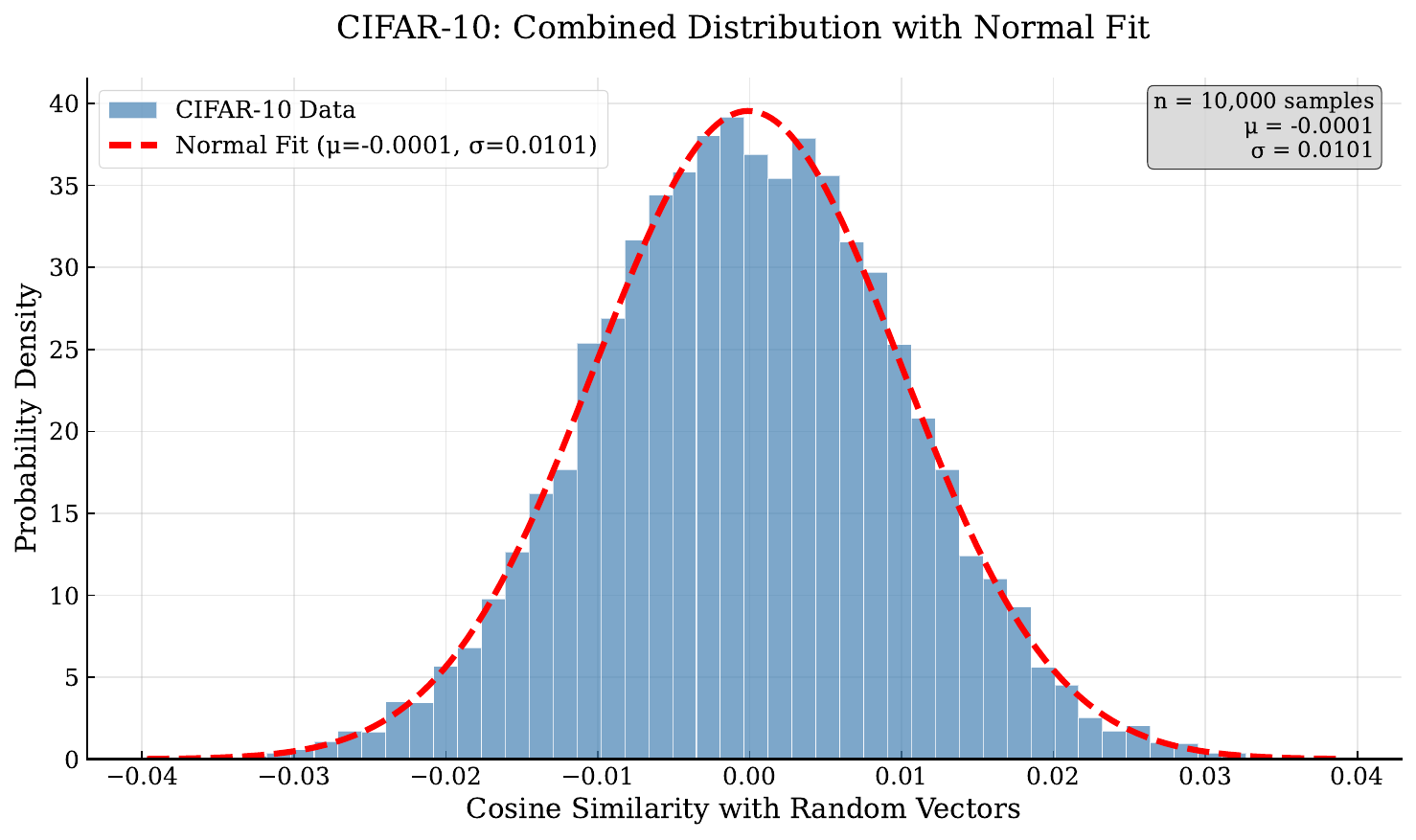}
  \caption{CIFAR-10}
  \label{fig:cifar-10}
\end{subfigure}
\hfill
\begin{subfigure}{0.33\textwidth}
  \centering
  \includegraphics[width=\linewidth]{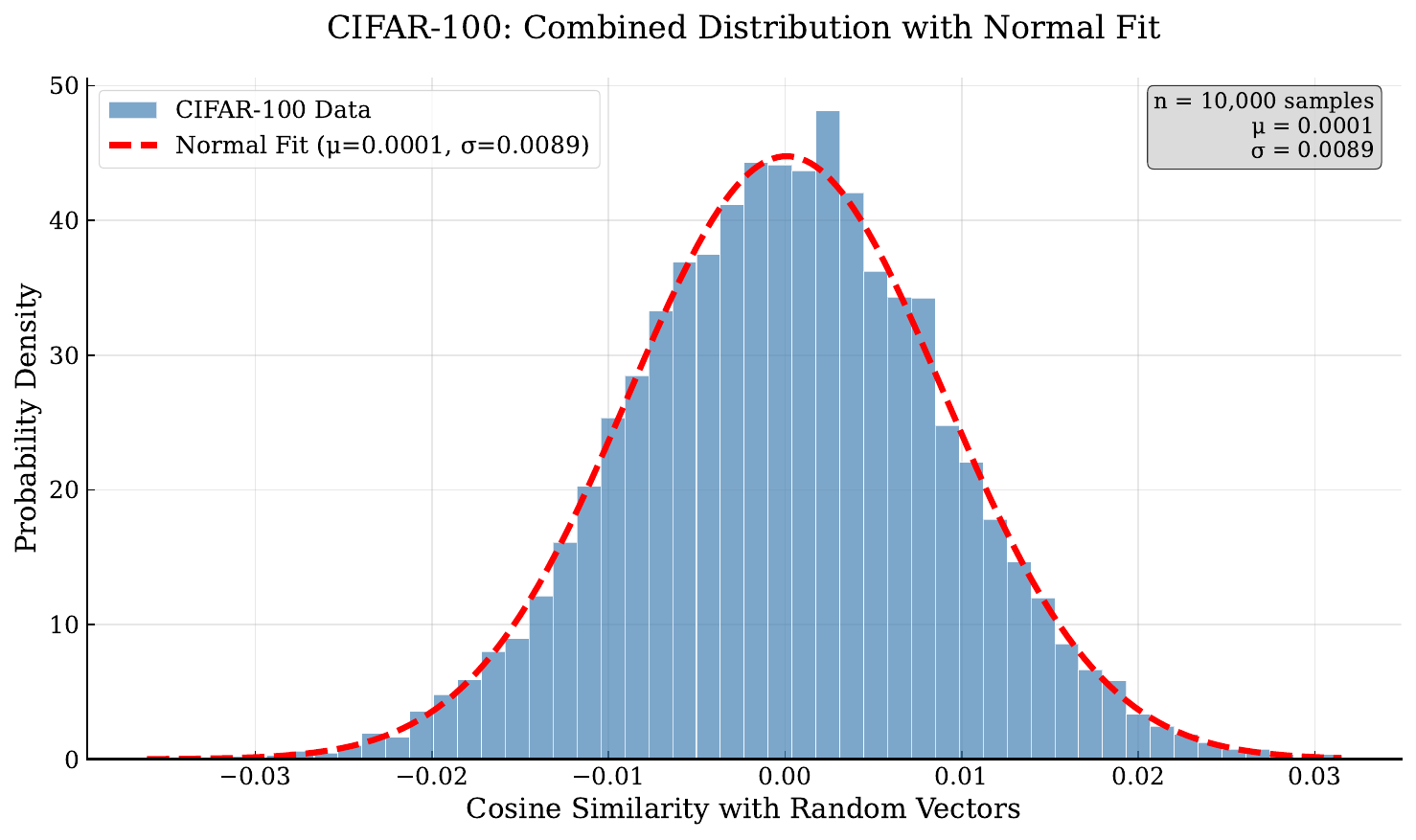}
  \caption{CIFAR-100}
  \label{fig:cifar-100}
\end{subfigure}
\hfill
\begin{subfigure}{0.33\textwidth}
  \centering
  \includegraphics[width=\linewidth]{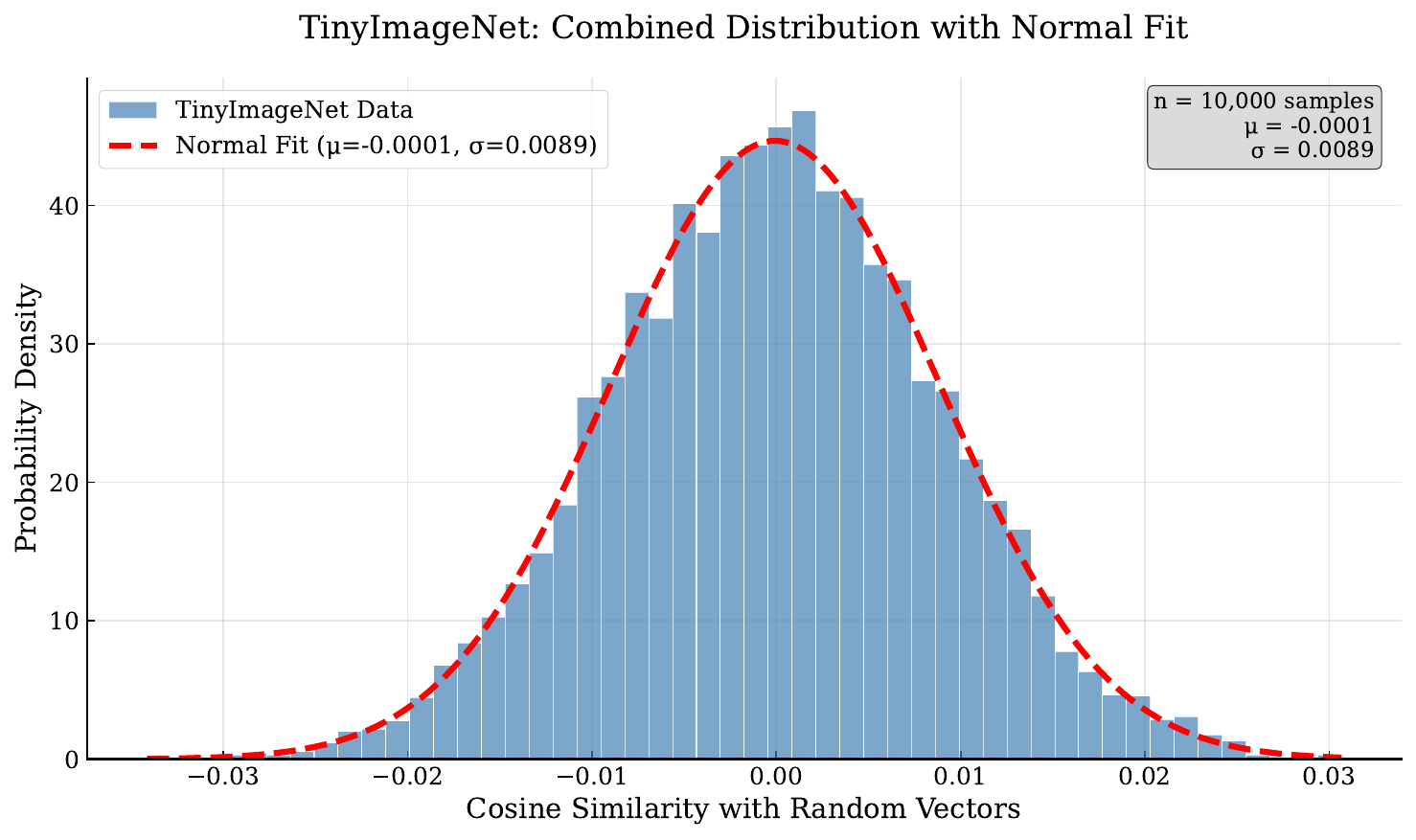}
  \caption{TinyImageNet}
  \label{fig:tiny_image}
\end{subfigure}
\caption{Combined cosine similarity distributions for different datasets on ResNet18 models, with fitted normal distributions.}
\label{fig:stat}
\end{figure*}

Given a suspect model $\theta_s$, we compute the cosine similarity between $\theta_s$ and $\tau$ and calculate the $z$-score
\begin{align}
z &= \frac{\cos(\theta_s, \tau) - \mu}{\sigma}
\end{align}
where $\mu$ and $\sigma$ denote the mean and standard deviation of cosine similarities between unwatermarked models and random vectors, which we empirically validate is approximately normal (see \Cref{fig:stat}). 
We consider a model $\theta_s$ watermarked if $z \geq 4$, which corresponds to a false positive rate of $\approx 3.2\cdot 10^{-5}$. 

\paragraph{Our method. }\Cref{alg:threshold_watermarking} implements our collaborative threshold watermarking protocol for FL.
In the setup phase, a trusted dealer or a DKG protocol assigns each of the $K$ clients a Shamir share $s_k$ of a secret watermark $\tau$, from which clients precompute public Lagrange coefficients $\lambda_k$ and embedding shares $w_k$.
In each training round $r$, the server broadcasts the current global model $\theta_{r-1}$, clients perform local training to obtain updates $\Delta\theta^{(k)}_r$ and track their magnitudes, and a secure aggregation step computes a global scaling factor $\text{scaletotal}_r$ that adaptively determines the watermark strength.
Each client embeds its watermark share into its local model update by submitting $u^{(k)}_r$ via secure aggregation, allowing the server to average the embedded updates into a new global model $\theta_r$.
After training completes, any coalition $S\subseteq\{1,\dots,K\}$ with $|S|\ge t$ can jointly verify the watermark on a suspect model $\theta_s$ by computing a verification statistic $z(\theta_s)$ from their shares optionally checking consistency with a public commitment and the watermark is accepted if $z(\theta_s)\ge z^\ast$.

Note that \textsc{LocalTrain}$(\theta,\mathcal{D}_k)$ denotes standard local training on client $k$'s dataset.
Any coalition of fewer than $t$ clients cannot reconstruct $\tau$ (by Shamir secrecy), and the only value revealed during verification is the final statistic $z$.
Moreover, with reconstruction-free verification, the same watermark key can be checked repeatedly across many checkpoints or models without revealing $\tau$ in the clear.

\begin{algorithm}[t]
    \caption{Threshold watermarking protocol.}
    \label{alg:threshold_watermarking}
    
    \begin{algorithmic}[1]
        \REQUIRE $K$ clients, threshold $t$, rounds $T$, strength $c$, EMA decay $\beta$, public evaluation points $\{x_k\}_{k=1}^K$
        \STATE \textbf{Setup:} Run \textsc{SetupTrustedDealer} (Alg.~\ref{alg:setup_dealer}) or dealer-free \textsc{SetupDKG} (Appendix~\ref{app:setup_dist:protocol}) to distribute Shamir shares $\{s_k\}_{k=1}^K$ of $\tau$ and publish nonce $\rho$ (and optional commitment $\mathcal{C}$).
        \STATE \textbf{Precompute:} Each client computes its public Lagrange coefficient $\lambda_k$ and embedding share $w_k \leftarrow \lambda_k s_k$
        \FOR{round $r = 1$ to $T$}
            \STATE Server broadcasts $\theta_{r-1}$
            \FOR{each client $k$ in parallel}
                \STATE $\theta_r^{(k)} \leftarrow \textsc{LocalTrain}(\theta_{r-1}, \mathcal{D}_k)$
                \STATE $\Delta_k \leftarrow \theta_r^{(k)} - \theta_{r-1}$ 
                \STATE $\text{ema}_{k} \leftarrow \beta\,\text{ema}_{k} + (1-\beta)\|\Delta_k\|_2$
                \STATE $\text{scale}_k \leftarrow c \cdot \|\Delta_k\|_2 \cdot \text{ema}_k$
            \ENDFOR
            \STATE $\text{scale}_{\text{total}} \leftarrow \textsc{SecAgg}(\{\text{scale}_k\}_{k=1}^K)$ and broadcast $\text{scale}_{\text{total}}$
            \STATE Each client forms $u_k \leftarrow \theta_r^{(k)} + \text{scale}_{\text{total}} w_k$ and contributes $u_k$ to \textsc{SecAgg}
            \STATE Server sets $\theta_r \leftarrow \frac{1}{K}\cdot \textsc{SecAgg}(\{u_k\}_{k=1}^K)$
        \ENDFOR
        \STATE \textbf{Verify:} Coalition $|\mathcal{S}|\ge t$ computes $z$ using its shares; optionally reconstructs $\tau$ to check commitment $(\rho,\mathcal{C})$
        \STATE Accept if $z\ge z^*$
    \end{algorithmic}
\end{algorithm}

%

%
    %
    %
    %

\section{Experiments}

\paragraph{Experimental Setup.} 
Experiments were conducted on the CIFAR-10, CIFAR-100 datasets \citep{krizhevsky2009learning} and Tiny ImageNet \citep{Le2015TinyIV}.
For image classification experiments, we use the ResNet-18 architecture~\citep{7780459}; all models were randomly initialized before training, and unless otherwise stated, results are averaged across three random seeds. 
We use FedAvg and vary the number of clients $K$ (4 to 128) while keeping the global batch size fixed at 2048. We run 300 rounds, each with one local epoch.
For CIFAR-10 and CIFAR-100, we hold out 20\% of the training data for validation and report test accuracy at the checkpoint with the highest validation accuracy.
To evaluate model utility, we report top-1 accuracy.
We quantify robustness by measuring the $z$-score after the attack.
We use a detection threshold $z^*=4$; we vary the watermark strength $c$ as indicated. 
Additionally, we report a language modeling
experiment with GPT-2 Small on WikiText-2.
We refer to Appendix \ref{app:implementation} for more implementation details.

\textbf{Baseline.} We compare against a naive per-client watermark baseline.
Each client $k$ samples an independent key $\tau_k$ and embeds it locally using the same update rule and scaling hyperparameter $c$ as our method. 
Verification for client $k$ uses the same one-sided $z$-test with key $\tau_k$. 
Under FedAvg aggregation, the watermark direction becomes proportional to $\frac{1}{K}\sum_{k=1}^K \tau_k$, whose expected norm shrinks as $1/\sqrt{K}$, explaining the loss of detectability as $K$ grows.

\subsection{Empirical Validation of Normality Assumption}
To validate the assumption of normality in the cosine similarity distribution (introduced in Section~\ref{sec:verification}), 
We trained five independent ResNet-18 models for each dataset. 
For each model $\theta_i$, we computed cosine similarities with 2000 random vectors. 
We then aggregated the similarities across models to obtain combined distributions from which $\mu$ and $\sigma$ were estimated.  As shown in Figure ~\ref{fig:stat}, individual model distributions exhibit consistent behavior, and the combined distributions closely match fitted normal curves. 

\subsection{Watermark Scalability}
\begin{figure*}[t]
\centering
\begin{subfigure}[t]{0.49\textwidth}
  \centering
  \includegraphics[width=\linewidth]{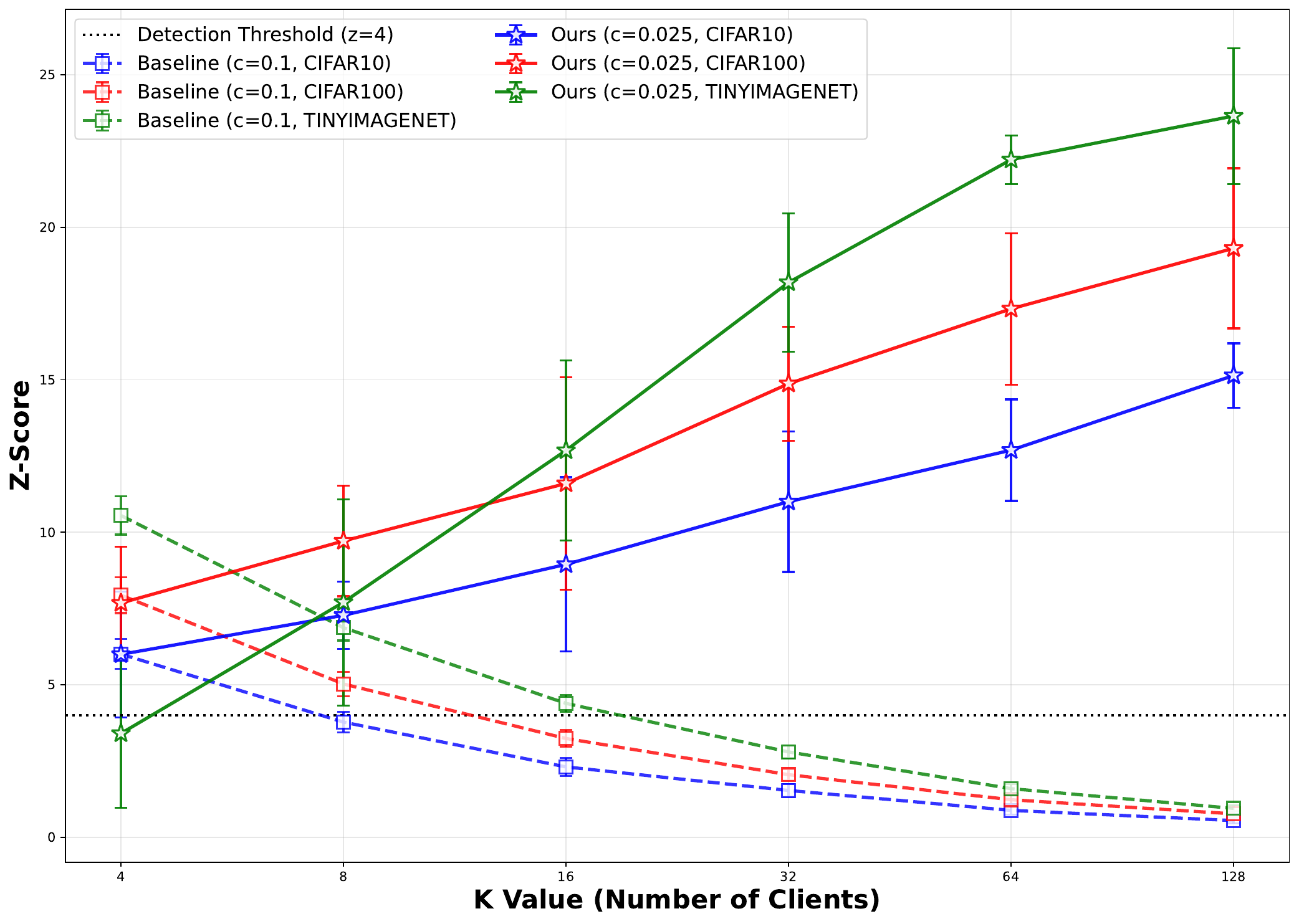}
  \caption{$z$-scores vs.\ number of clients}
  \label{fig:baseline_vs_ours}
\end{subfigure}
\hfill
\begin{subfigure}[t]{0.49\textwidth}
  \centering
  \includegraphics[width=\linewidth]{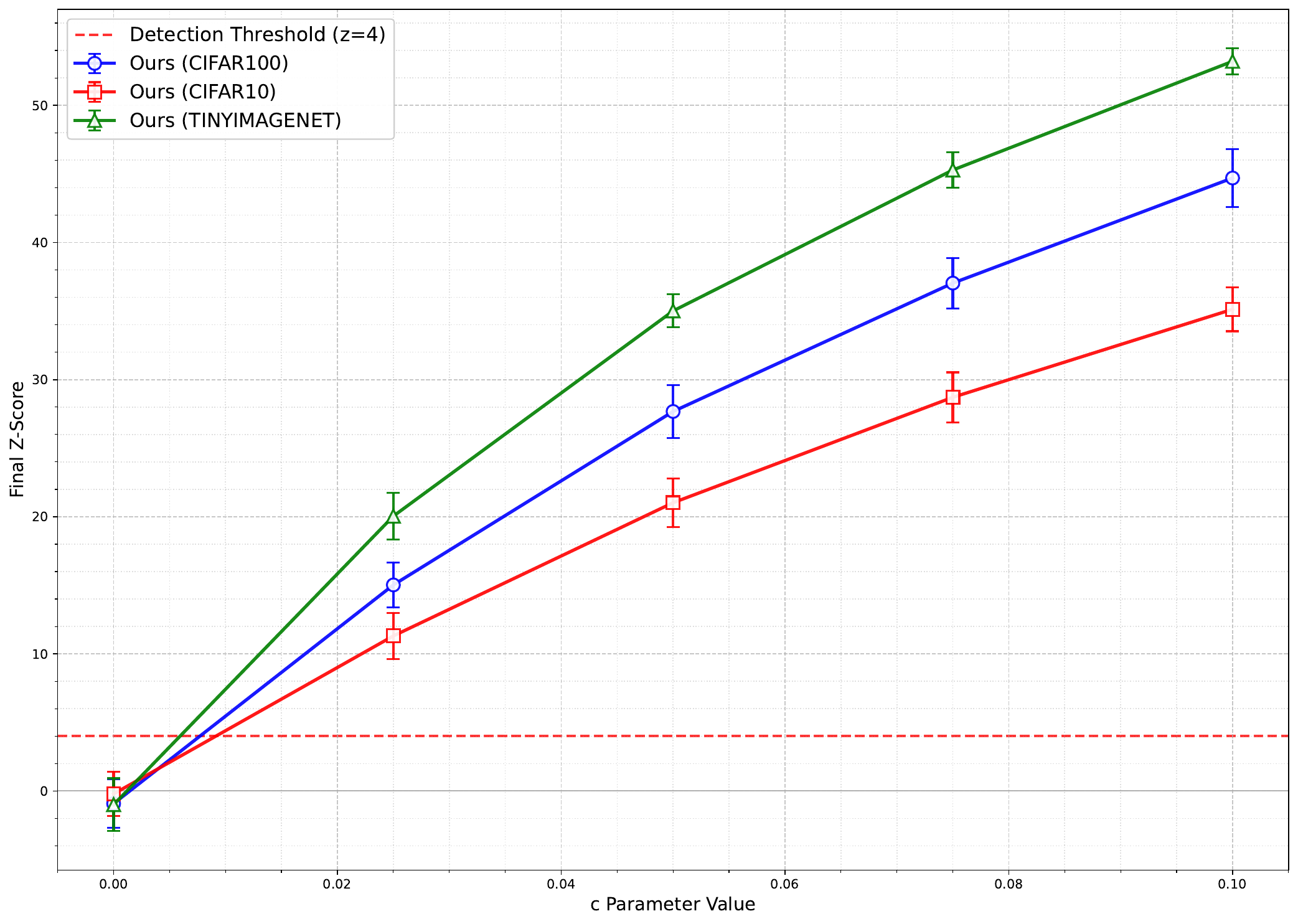}
  \caption{$z$-scores vs.\ $c$ with $K=32$.}
  \label{fig:z_vs_c}
\end{subfigure}
\caption{(a) Our method sustains statistically significant $z$-scores up to $K=128$, whereas the baseline collapses beyond $K=16$. 
(b) Increasing the scaling factor $c$ consistently boosts $z$-scores across datasets, showing that watermark strength is tunable.}
\label{Comparison}
\end{figure*}

\Cref{fig:baseline_vs_ours} illustrates the effect of scaling the number of clients on the watermark signal strength for both the baseline and our proposed method. 
We observe that for the baseline, each client embeds a unique signal, which causes the overall signal to weaken as the number of clients increases.
Notably, when $K \geq 16$, the $z$-score is not statistically significant and falls below our detection threshold for the baseline.
In contrast, our collaborative watermarking approach supports scaling to larger $K$ (demonstrated up to $K=128$) even with a smaller scaling factor ($c=0.025$, compared to $c=0.1$ in the baseline). 
This comparison is conservative for our method.
We give the baseline a larger $c$ to improve its detectability at small $K$, but it still diminishes as $K$ grows. 
Figure~\ref{fig:z_vs_c} and Table~\ref{tab:acc_vs_c} enable matched-utility comparisons by showing how $z$ and accuracy vary with $c$.

Figure~\ref{fig:z_vs_c} shows that the statistical significance with which our watermark can be detected increases predictably with the watermark strength hyperparameter~$c$ (scaling factor). 
As~$c$ increases, the corresponding $z$-scores rise consistently across all datasets.
which shows that our scheme is both scalable and tunable by adjusting $c$ to control the watermarking strength. 
In the next section, we further investigate the trade-off between watermark strength and model accuracy.

\subsection{Watermark Fidelity}
Table \ref{tab:acc_vs_c} shows the model's accuracy relative to the watermark embedding strength $c$.
At low watermark strengths, the impact on accuracy is small. 
For example, at $c=0.025$, test accuracy drops by only 0.12 percentage points (pp) on CIFAR-10, 0.21 pp on CIFAR-100, and 0.25 pp on Tiny ImageNet.
Increasing $c$ to 0.05 or 0.075 yields slightly larger but still modest reductions.
CIFAR-10 drops by 0.28 pp and 0.64 pp, respectively. 
CIFAR-100 drops by 1.15 pp and 2.13 pp, and Tiny ImageNet by 0.47 pp and 0.57 pp.
This demonstrates that even at moderate strengths, the model maintains high accuracy.

At higher watermark strengths (e.g., $c=0.1$), accuracy noticeably degrades: 1.1 pp for CIFAR-10, 4.6 pp for CIFAR-100, and 2.5 pp for Tiny ImageNet.
Overall, $c=0.025$ provides a strong default trade-off in our experiments, preserving accuracy while yielding reliable detectability.

\begin{table}[ht]
\centering
\caption{Test accuracy (\%) across datasets for different watermark strengths $c$ ($K=32$).}
\label{tab:acc_vs_c}
\setlength{\tabcolsep}{1.5pt} 
\begin{tabular}{c|c|c|c}
\toprule
$c$ & CIFAR-10 & CIFAR-100 & TinyImageNet \\
\midrule
No Watermark & 88.08 $\pm$ 0.53 & 61.33 $\pm$ 1.56 & 53.98 $\pm$ 0.88\\
0.025 & 87.93 $\pm$ 0.65 & 61.31 $\pm$ 0.78 & 53.67 $\pm$ 0.52\\
0.050 & 87.62 $\pm$ 0.49 & 60.47 $\pm$ 0.58 & 53.51 $\pm$ 0.75\\
0.075 & 87.36 $\pm$ 0.13 & 59.55 $\pm$ 0.98 & 53.45 $\pm$ 0.68\\
0.100 & 86.89 $\pm$ 0.32 & 58.80 $\pm$ 1.08 & 52.59 $\pm$ 0.49\\
\bottomrule
\end{tabular}
\end{table}

\subsection{Non-IID Data Heterogeneity}

We next evaluate 
watermark detectability under statistical heterogeneity. We use the setup under a balanced Dirichlet label-skew partition with
$K=32$ and $c=0.025$. All clients receive the same number of training examples,
while class proportions are sampled from a Dirichlet distribution with concentration
parameter $\gamma$. Smaller $\gamma$ corresponds to stronger label skew and thus
greater statistical heterogeneity.

Table~\ref{tab:noniid} reports the results for
$\gamma \in \{0.25, 0.5, 0.75, 1.0\}$. Across all tested values of $\gamma$, the
watermark remains highly detectable, with z-scores between $12.58$ and $12.72$,
well above the detection threshold $z^\ast = 4$. The utility gap relative to the non-watermarked model also remains small across all settings. These results suggest
that statistical heterogeneity has little effect on the watermark
detectability.
\begin{table}[ht]
\centering
\caption{CIFAR-100 under balanced Dirichlet label skew ($K=32$).}
\label{tab:noniid}
\setlength{\tabcolsep}{3pt}
\begin{tabular}{c|c|c|c}
\toprule
$\gamma$ & Z-score & Non-WM Acc. & WM Acc. ($c=0.025$) \\
\midrule
0.25 & 12.72 $\pm$ 1.61 & 60.18 $\pm$ 0.55 & 60.02 $\pm$ 0.28 \\
0.50 & 12.60 $\pm$ 2.72 & 61.37 $\pm$ 0.29 & 61.11 $\pm$ 0.59 \\
0.75 & 12.58 $\pm$ 2.85 & 61.16 $\pm$ 0.31 & 60.86 $\pm$ 0.44 \\
1.00 & 12.60 $\pm$ 2.88 & 61.31 $\pm$ 0.11 & 60.78 $\pm$ 0.07 \\
\bottomrule
\end{tabular}
\end{table}

\subsection{Language Modeling}

We further evaluate the method beyond image classification tasks.
Table~\ref{tab:gpt2} summarizes results for GPT-2 Small on WikiText-2.
Without watermarking, the model achieves perplexity $10.52 \pm 0.05$ and
z-score $0.17 \pm 1.69$, consistent with the absence of a watermark signal.
Once watermarking is enabled, detectability remains strong: for
$c \in \{0.50, 0.75, 1.00\}$, the z-score stays well above the detection
threshold. At the same time, the increase in perplexity is small, indicating
that the impact on model utility remains limited in this setting.

\begin{table}[h]
\centering
\caption{GPT-2 Small on WikiText-2: perplexity and corresponding z-score ($k=4$).}
\label{tab:gpt2}
\setlength{\tabcolsep}{3pt}
\begin{tabular}{c|c|c}
\toprule
$c$ & Perplexity & Z-score \\
\midrule
No Watermark & 10.52 $\pm$ 0.05 & 0.17 $\pm$ 1.69 \\
0.50 & 10.58 $\pm$ 0.03 & 6.89 $\pm$ 2.51 \\
0.75 & 10.63 $\pm$ 0.06 & 8.13 $\pm$ 1.75 \\
1.00 & 10.70 $\pm$ 0.07 & 9.63 $\pm$ 1.30 \\
\bottomrule
\end{tabular}
\end{table}

\subsection{Watermark Robustness}
%
%

We evaluate robustness against common post-training removal attacks and adaptive attackers, following prior model-watermarking work \citep{10.1145/3078971.3078974,10.5555/3277203.3277324,pmlr-v162-bansal22a}.
\begin{enumerate}[itemsep=0pt, topsep=0pt]
    \item \textbf{Fine-Tuning:} Fine-tune the released model for 100 epochs using AdamW ($1\times10^{-3}$ learning rate, $1\times10^{-4}$ weight decay, batch size 128) on $p\% \in \{1,5,10,20\}$ of the training data. 
    \item \textbf{Quantization:} Weight-only quantization of Conv/Linear layers of the final model checkpoint.  
    \item \textbf{Pruning:} Magnitude pruning (global unstructured) or structured channel pruning with ratios in $\{0.3,0.5,0.7,0.9\}$.
    \item \textbf{Distillation:} Train a student network for 100 epochs (Adam, $1\times10^{-3}$, batch size 128) on $p\%$ of the training data with temperature $T=3$ and loss weight $\alpha=0.5$.
    \item \textbf{Adaptive Fine-Tuning:} Use intermediate global checkpoints to estimate a watermark direction and fine-tune to reduce alignment with the estimate (same data/epoch budgets).
\end{enumerate}

\begin{figure*}[t]
  \centering
  \includegraphics[width=\textwidth]{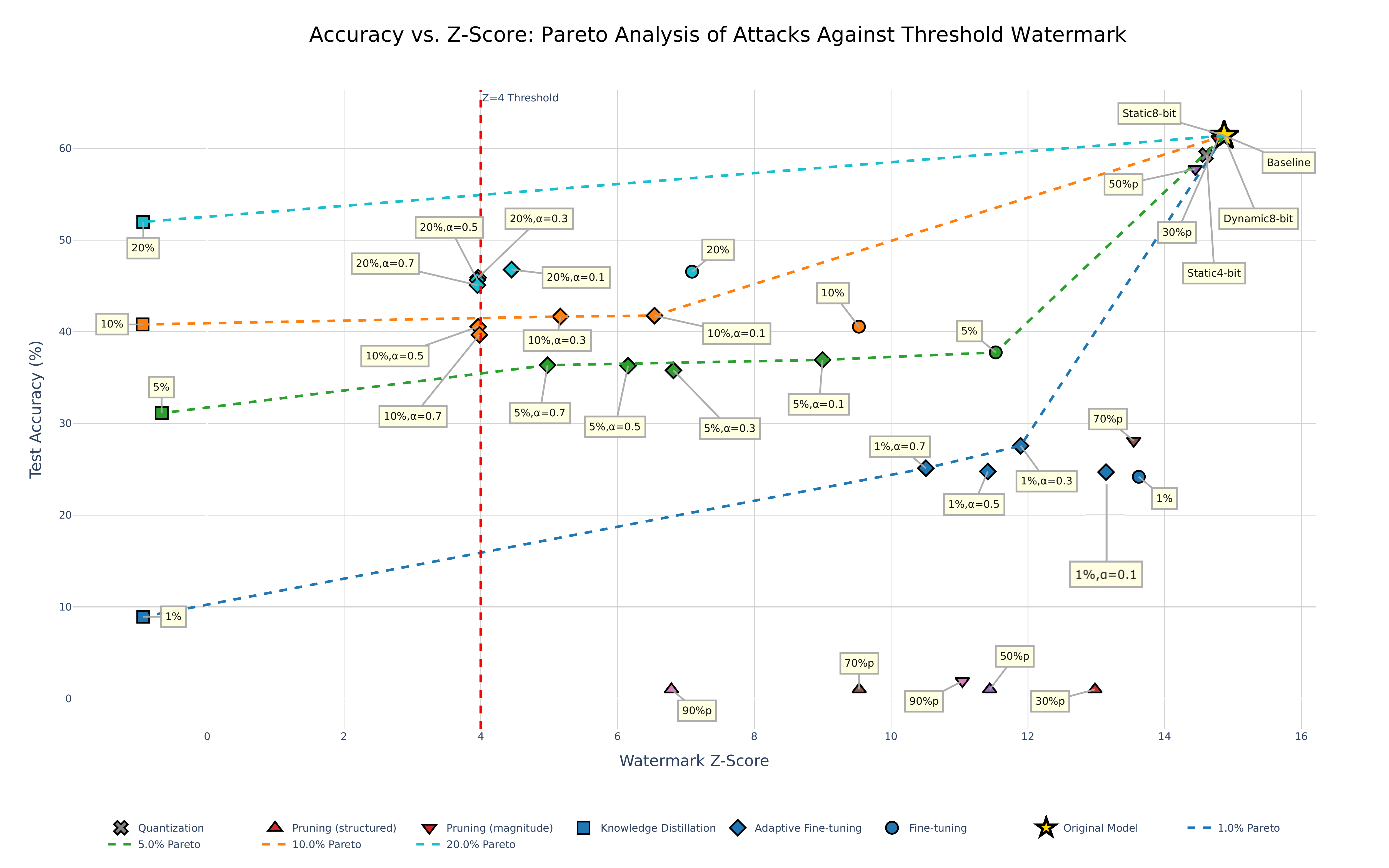}
  \caption{Robustness analysis on CIFAR-100 with $K=32$ and $c=0.025$. 
  We report the trade-off between task accuracy and watermark $z$-score under five attack types: (i) adaptive fine-tuning, (ii) plain fine-tuning, (iii) knowledge distillation, (iv) pruning (magnitude and structured), and (v) quantization. The original model is shown as a star. Dashed curves denote Pareto frontiers for 1\%, 5\%, 10\%, and 20\% of the training data, while the red dashed line marks the detection threshold ($z=4$).}
  \label{fig:robustness_cifar100}
\end{figure*}

\paragraph{Results.}
Figure~\ref{fig:robustness_cifar100} shows the robustness results on CIFAR-100 ($K=32$, $c=0.025$) via Pareto frontiers. 
We observe that even with larger attack budgets (up to 20\% of the data) or structural modifications such as 90\% pruning, watermark $z$-scores remain above the detection threshold ($z=4$). 
Adaptive fine-tuning yields better results for the attacker, but it still cannot erase the watermark without substantial accuracy degradation.
Figure~\ref{fig:robustness_cifar100} plots all post-attack checkpoints as \((\text{accuracy}, z)\) pairs.
Points above the dashed line mean that the watermark remained detectable. 
The dashed Pareto curves summarize the best measured trade-offs at each data budget. 
With fine-tuning, pruning, and quantization, the watermark remains detectable until the attacker incurs a significant drop in accuracy. 

The only attack that reliably removes our watermark is distillation, as expected, since we use a white-box watermark. 
However, distillation requires (i) a high computational effort to re-train the model and (ii) substantial amounts of training data.   
We highlight that our watermark is not robust when the adversary has sufficient training data (see the top-left attack in Figure~\ref{fig:robustness_cifar100}). This is well known and agrees with previous works~\citep{10.3389/fdata.2021.729663,lukas2021sokrobustimageclassification}.
We refer to Appendix \ref{app:robustness} for more details on robustness.

\section{Discussion }

We propose threshold watermarking as a problem to be studied, especially as LLMs are being trained using FL ~\cite{sani2024futurelargelanguagemodel}. Our work targets shared model ownership in FL, ensuring that no single participant can unilaterally remove a watermark.
The core technical contribution is a protocol that combines (i) threshold secret sharing of the watermark key for $(t,K)$-controlled verification, (ii) distributed watermark embedding under secure aggregation with an untrusted server, and (iii) empirical evaluation of fidelity and robustness at scale. By enforcing a shared embedding direction, our method avoids the detectability loss observed in per-client watermarking baselines as $ K$ increases, where individual contributions diminish. 
Our results show that even with access to the full training trajectory, watermark removal remains difficult under realistic attack budgets. (iv) Experiments under balanced Dirichlet label skew and on GPT-2
Small  suggests that the method is not confined to the default IID
image-classification setting, where in both cases, watermark detectability remains
strong while utility degradation stays limited.

\paragraph{Limitations.} 
Our current instantiation is white-box, meaning that \textsc{Verify} assumes access to model parameters.
We assume honest-but-curious participants and do not handle training-time adversaries (e.g., Byzantine clients or a malicious server that deviates from the protocol). 
Extending threshold watermarking to these settings would require additional security and robustness mechanisms. 
While our experimental evaluation primarily targets image classification, we further assess our method in a GPT-2 Small language model fine-tuning setting. A comprehensive evaluation of larger language models is left for future work.
%
%
Finally, since the global model trajectory is observable, an attacker may attempt to estimate the watermark key from checkpoints.
We empirically instantiate such adaptive removal attacks with substantial auxiliary knowledge, but do not provide a theoretical lower bound on the difficulty of estimating the key; thus, stronger adaptive attacks may exist. 

\section{Related Work}

In centralized training, watermarking embeds a secret signal that later proves ownership of a model~\citep{10.3389/fdata.2021.729663,lukas2021sokrobustimageclassification}. 
Existing schemes span white-box approaches that encode signals directly in parameters~\citep{10038500,10.1145/3297858.3304051,10.1145/3442381.3450000,doi:10.2352/ISSN.2470-1173.2020.4.MWSF-022} and black-box approaches that verify via trigger queries~\citep{10.5555/3277203.3277324,272262,pmlr-v162-bansal22a,yang2023fedzkpfederatedmodelownership}.

In FL, watermarking must account for distributed ownership and limited trust.
Secret keys should not be held by a single party, and embedding should remain robust even when the server is untrusted (see \citet{lansari2023federatedlearningmeetswatermarking} for a survey). 
Prior work typically assumes either a trusted server that embeds centrally~\citep{9603498,10504977,9859395} or designated clients that embed via local trigger data~\citep{9847383,10.1145/3630636,9658998,10.1145/3651671.3651710,liang2023fedcipfederatedclientintellectual,xu2024robwerobustwatermarkembedding}.
Our focus is complementary, since we enable a $(t,K)$-threshold verification via Shamir secret sharing and enable distributed embedding under secure aggregation.

\section{Conclusion}
We proposed $(t, K)$-threshold watermarking for FL, enabling collaborative model ownership without requiring a trusted server.
Our construction guarantees that only coalitions of at least $t$ clients can verify provenance, while smaller groups learn nothing beyond the protocol’s output. 
We instantiated our protocol in the \textit{white-box setting} and empirically demonstrated its robustness against watermark-removal attacks, including adaptive attackers. 
Our results show that threshold watermarking is practical and scalable to large $K$, making it suitable for collaborative ownership of machine learning models.
More broadly, we hope this work supports large-scale model training and enables reliable attribution through collaborative watermarking.

\section*{Impact Statement}

This paper introduces a threshold watermarking protocol for federated
learning that enables shared ownership and collective provenance
verification of jointly trained models. By requiring a coalition of
clients to verify a watermark, the method supports collaborative
settings in which no single participant should unilaterally claim
ownership or remove attribution. This is particularly relevant for
applications such as healthcare or cross-organizational model training,
where data and compute are contributed by multiple parties.

\bibliography{example_paper}
\bibliographystyle{icml2026}

\newpage
\appendix
\onecolumn
\section{Notation Table}
\begin{table}[H]
\centering
\caption{Notation table for collaborative threshold watermarking.}
\label{tab:notation}
\begin{tabular}{ll}
\toprule
\textbf{Symbol} & \textbf{Definition} \\
\midrule
$K$ & Total number of clients in federated learning (FL). \\
$k$ & Index of a client, $k \in \{1,\dots,K\}$. \\
$D_k$ & Local dataset of client $k$. \\
$n_k$ & Number of samples in $D_k$. \\
$n$ & Total number of samples, $n = \sum_{k=1}^K n_k$. \\
$F_k(\theta)$ & Local loss function of client $k$. \\
$F(\theta)$ & Global objective, weighted sum of local losses. \\
\midrule
$\theta_r$ & Global model parameters at round $r$. \\
$\theta_0$ & Initial global model. \\
$\theta_w$ & Final watermarked global model after $T$ rounds. \\
$\theta^{(k)}_r$ & Locally trained model of client $k$ at round $r$. \\
$\nabla \theta^{(k)}_r$ & Local model update of client $k$ at round $r$. \\
$\theta_s$ & Suspect model under verification. \\
\midrule
$\tau$ & Secret watermarking key vector. \\
$\tau'$ & Estimated key (attack approximation of $\tau$). \\
$\mathcal{C}$ & Public commitment to $\tau$ (from dealer). \\
$\rho$ & Public nonce (randomness) used in the commitment $\mathcal{C}$\\
$s_i$ & Shamir secret share of $\tau$ held by client $i$. \\
$w_i$ & Additive share of $\tau$ (trustless-server setting). \\
$S_r$ & Set of active clients in round $r$ (participating clients). \\
$a_{k,r}$ & Normalized aggregation weight of client $k \in S_r$ in round $r$, with $\sum_{k \in S_r} a_{k,r} = 1$. \\
$\tilde{w}_{k,r}$ & Weighted embedding share of client $k$ in round $r$, defined as $\tilde{w}_{k,r} = a_{k,r}^{-1} w_k$. \\
\midrule
$c$ & Scaling constant for watermark strength. \\
$\beta$ & EMA decay factor for update magnitudes. \\
$\text{ema}_k$ & Exponential moving average tracker for client $k$. \\
$\text{scale}_k$ & Scaling factor for client $k$’s watermark perturbation. \\
$\text{scale}_{\text{total}}$ & Aggregated watermark scaling factor. \\
\midrule
$z$ & Standardized verification statistic. \\
$z^{*}$ & Detection threshold for watermark verification. \\
$\mu, \sigma$ & Mean and standard deviation of cosine similarities (unwatermarked models). \\
\midrule
$E$ & Local training epochs per round. \\
$T$ & Number of global FL rounds. \\
$\eta$ & Learning rate. \\
\midrule
$z_s, z_t$ & Student and teacher logits (distillation attack). \\
$T_{\text{distill}}$ & Temperature parameter for distillation softening. \\
$\alpha$ & Mixing factor (attack or distillation trade-off). \\
$p\%$ & Fraction of training data available to adversary during attacks. \\
\bottomrule
\end{tabular}

\end{table}

\section{Dealer-Free Setup (DKG)}
\label{app:setup_dist}

If a trusted dealer is unavailable, clients can generate shares of a random watermark key without any single party learning $\tau$ in the clear using a standard distributed key generation (DKG) protocol.
At a high level, each client contributes randomness, and the resulting secret key is implicitly defined as the sum of all contributions.

\subsection{Dealer-Free Setup Protocol}
\label{app:setup_dist:protocol}

Algorithm~\ref{alg:setup_dealer_dist} describes a dealer-free setup based on Shamir secret sharing.
Each client $k$ samples an additive share $w_k$ and constructs a degree-$(t-1)$ polynomial $P_k$ with constant term $P_k(0)=w_k$.
Clients privately exchange polynomial evaluations and locally aggregate them to obtain Shamir shares $s_i$ of the implicit secret $\tau=\sum_{k=1}^K w_k$.
No single client ever reconstructs $\tau$, and Shamir’s $(t,K)$-threshold secrecy is preserved.

\begin{algorithm}
    \caption{Dealer-free setup via distributed key generation (DKG).}
    \label{alg:setup_dealer_dist}
    \begin{algorithmic}[1]
        \REQUIRE $K$ clients, threshold $t$
        \STATE Publicly sample nonce $\rho$
        \FOR{each client $k$ in parallel}
            \STATE Sample additive share $w_k \sim \mathcal{N}(0,I_d/K)$
            \STATE Sample degree-$(t-1)$ polynomial $P_k$ with $P_k(0)=w_k$
            \FOR{each client $i \in \{1,\dots,K\}$}
                \STATE Privately send $P_k(i)$ to client $i$
            \ENDFOR
        \ENDFOR
        \FOR{each client $i$}
            \STATE Set Shamir share $s_i \leftarrow \sum_{k=1}^K P_k(i)$
            \STATE Keep additive share $w_i$
        \ENDFOR
    \end{algorithmic}
\end{algorithm}

\subsection{Dealer-Free Setup Overhead}
\label{app:setup_dist:overhead}

\begin{figure*}
  \centering
  \includegraphics[width=\textwidth]{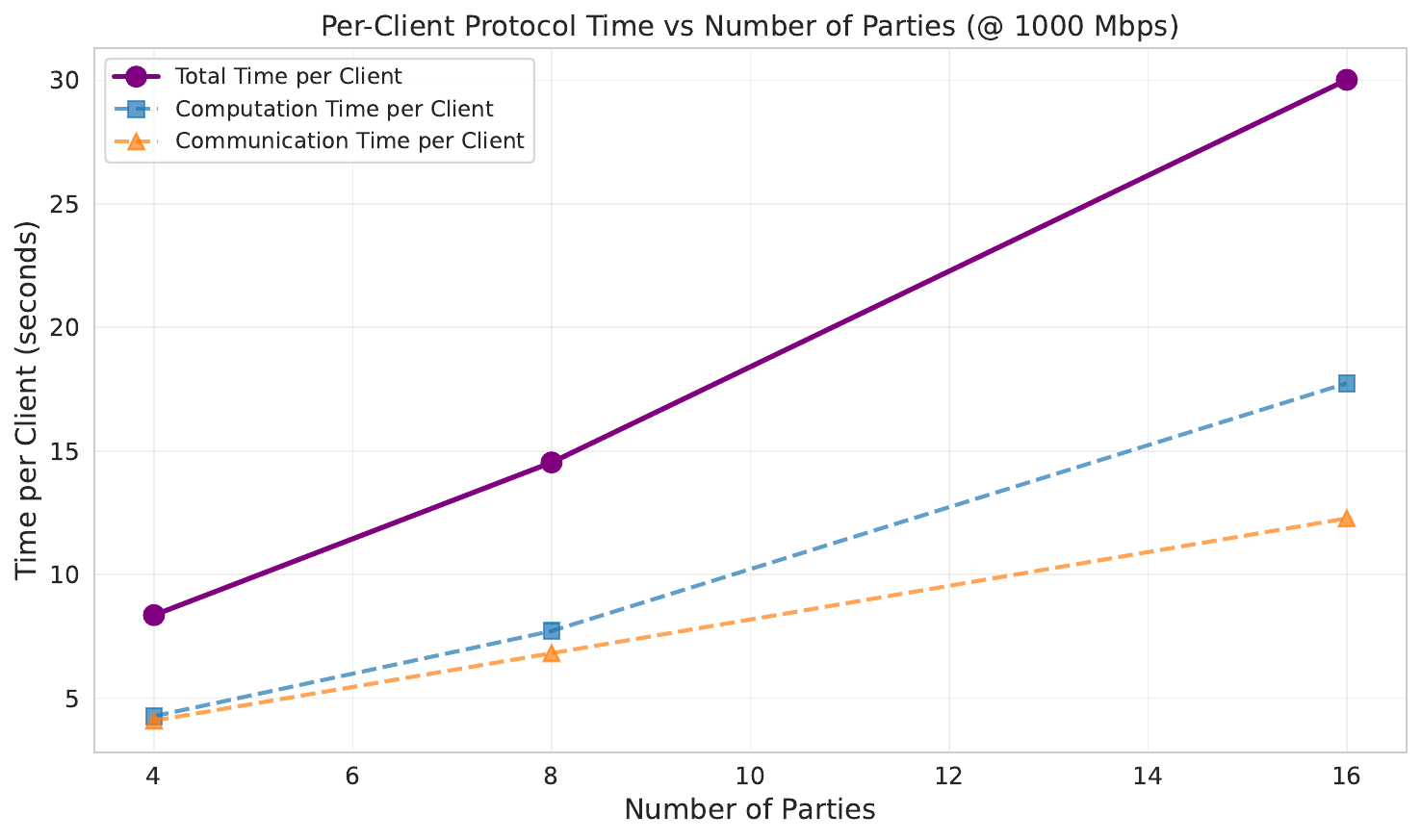}
  \caption{Per-client DKG protocol runtime versus number of participating clients under a 1000~Mbps network, showing total time and the breakdown between computation and communication.}
  \label{fig:per_client_time}
\end{figure*}

Figure~\ref{fig:per_client_time} reports the per-client runtime of the DKG protocol as a function of the number of participating clients $K$, measured under a simulated 1000~Mbps network.
We decompose the total runtime into computation and communication costs.
The observed scaling matches the theoretical complexity: for fixed model dimension $d$, per-client communication grows as $O(Kd)$ and per-client computation as $O(Ktd)$, consistent with Shamir-based DKG.
At the system level, this corresponds to an overall communication complexity of $O(K^2 d)$ and a total computation complexity of $O(K^2 t d)$.
As $K$ increases, the threshold $t$ grows proportionally, leading to computation dominating communication, which explains the widening gap at larger scales.

\section{Implementation Details}
\label{app:implementation}

\textbf{Datasets.} We conduct experiments on CIFAR-10, CIFAR-100 \citep{krizhevsky2009learning}, and Tiny ImageNet \citep{Le2015TinyIV}. Since CIFAR-10 and CIFAR-100 do not provide validation sets, we split the original training set into 80\% for training and 20\% for validation. For evaluation, we select the model checkpoint with the highest validation accuracy. For CIFAR-10 and CIFAR-100, we apply the AutoAugment policy designed for CIFAR-10. For Tiny ImageNet, training images are augmented using random horizontal flips ($p=0.5$), random rotations ($\pm15^\circ$), and color jittering (brightness, contrast, and saturation up to 0.4; hue up to 0.1). For the non-IID CIFAR-100 experiment, we use a balanced
Dirichlet label-skew partition in which all clients receive the same
number of training examples, while class proportions are sampled from a
Dirichlet distribution with concentration parameter
$\gamma \in \{0.25, 0.5, 0.75, 1.0\}$.

\textbf{Model.} All experiments use the ResNet-18 architecture \citep{7780459}, with models initialized randomly before training. Implementations and training are carried out in PyTorch \citep{10.5555/3454287.3455008}.

\textbf{Training.} In our FL setup, we vary the number of clients from 4 to 128 while keeping the global batch size fixed at 2048, ensuring an equal number of batches across experiments. We use the AdamW optimizer \citep{loshchilov2019decoupled} with a learning rate of $1\times10^{-3}$, weight decay of $1\times10^{-4}$, and betas $(0.9,0.999)$. Training runs for 300 rounds with a single epoch per round. The EMA decay factor $\beta$ for our watermark is set to 0.9. To ensure reproducibility and reduce the risk of seed overfitting, all experiments are repeated with three random seeds: 0, 1, and 2.

\textbf{Language modeling.} We additionally evaluate GPT-2 Small on WikiText-2 in a
federated fine-tuning setting. We use $K=4$ clients, train for 20 federated
rounds with AdamW, and use a learning rate of $1 \times 10^{-4}$. For
evaluation, we retain the checkpoint with the lowest perplexity.

\textbf{Hardware.} All training and post-training attack experiments are conducted on a single NVIDIA RTX A6000 GPU (48 GB VRAM).

\section{Robustness Analysis}
\label{app:robustness}

This section details the robustness evaluation of our watermarking method. We describe the attacks used to showcase our watermark robustness, along with the experimental setup, so that our results can be reproduced and extended by other researchers. Robustness is assessed under a range of attack scenarios, with measurements reported in terms of (i) classification accuracy on the test set and (ii) watermark detectability via the standardized $z$-score.

\subsection{Fine-Tuning}
Fine-tuning is the most direct strategy an attacker can attempt, simply retrain the released model on a subset of data in the hope of diminishing the watermark signal.
 
The attack is assumed to have access to only a subset of the training data, with size $p\% \in \{1,5,10,20\}$ of the full training dataset. Each subset is sampled uniformly at random from the training set with fixed seeds to ensure reproducibility. Fine-tuning proceeds for 100 epochs using the AdamW optimizer with learning rate $1\times10^{-3}$, weight decay $1\times10^{-4}$, betas $(0.9,0.999)$, and batch size $128$. 

\begin{figure*}[h]
\centering
\begin{subfigure}[t]{0.6\textwidth}
  \centering
  \includegraphics[width=\linewidth]{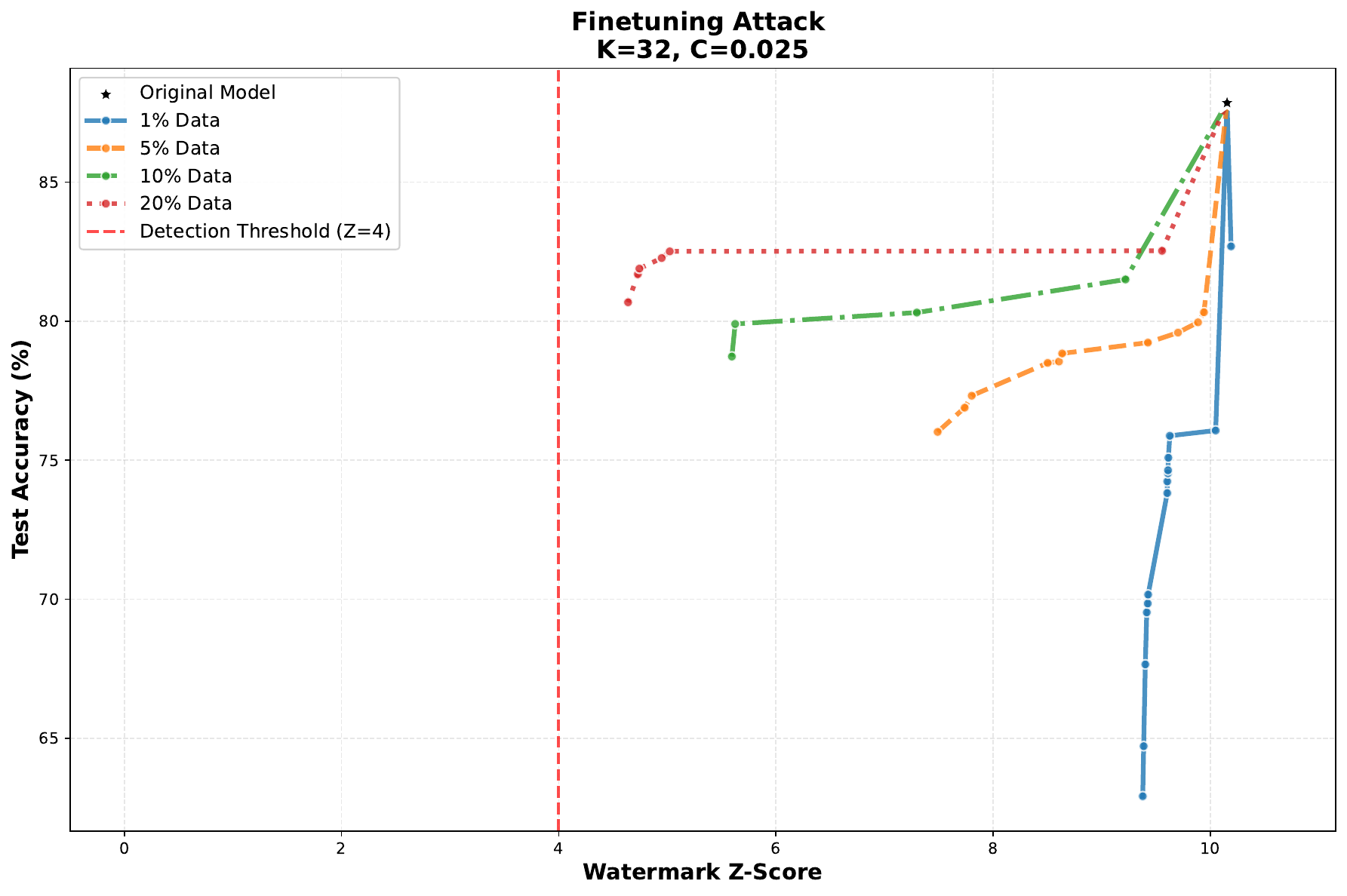}
  \caption{CIFAR-10}
  \label{fig:finetune_cifar10}
\end{subfigure}
\hfill
\begin{subfigure}[t]{0.6\textwidth}
  \centering
  \includegraphics[width=\linewidth]{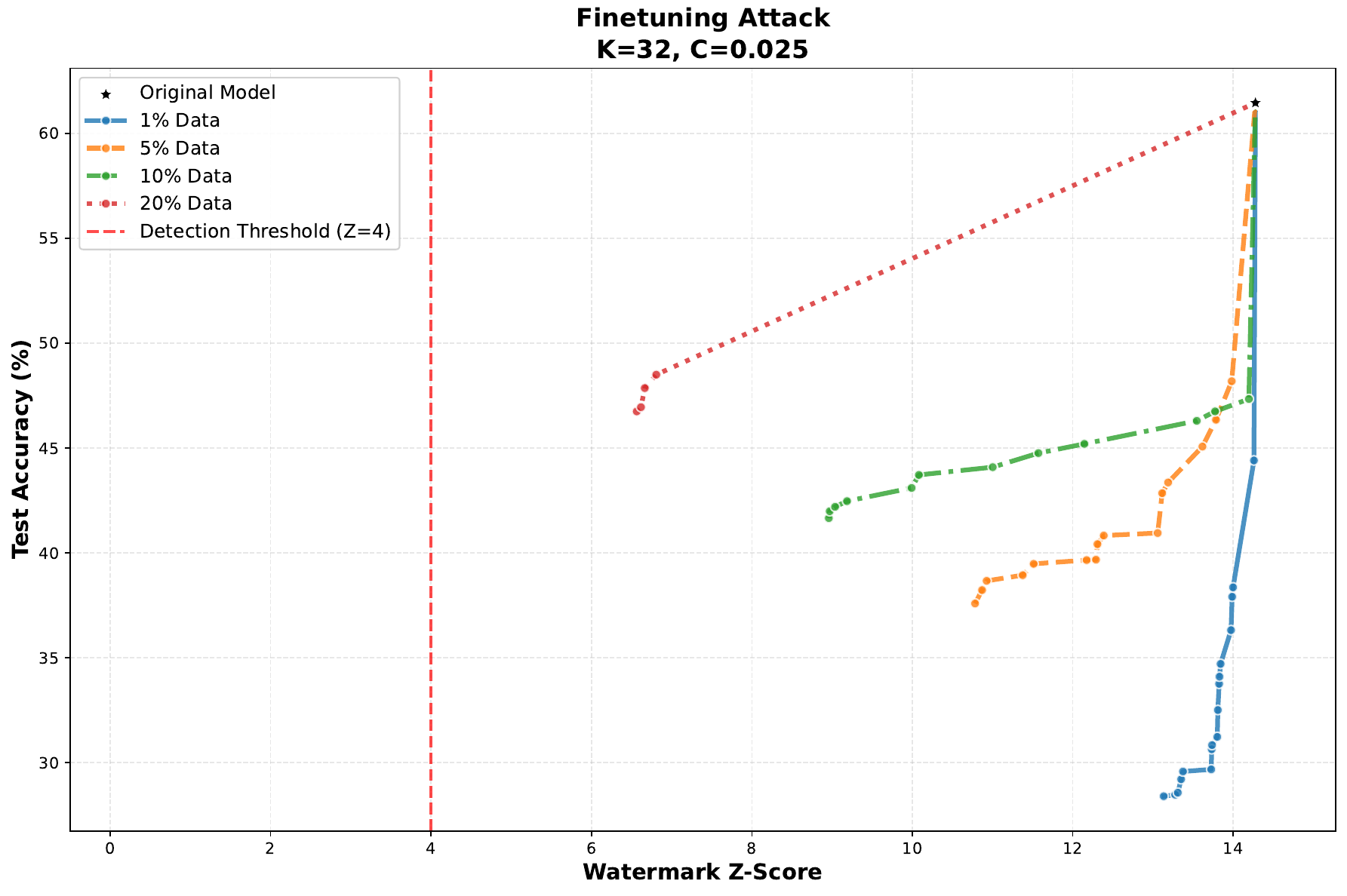}
  \caption{CIFAR-100}
  \label{fig:finetune_cifar100}
\end{subfigure}
\hfill
\begin{subfigure}[t]{0.6\textwidth}
  \centering
  \includegraphics[width=\linewidth]{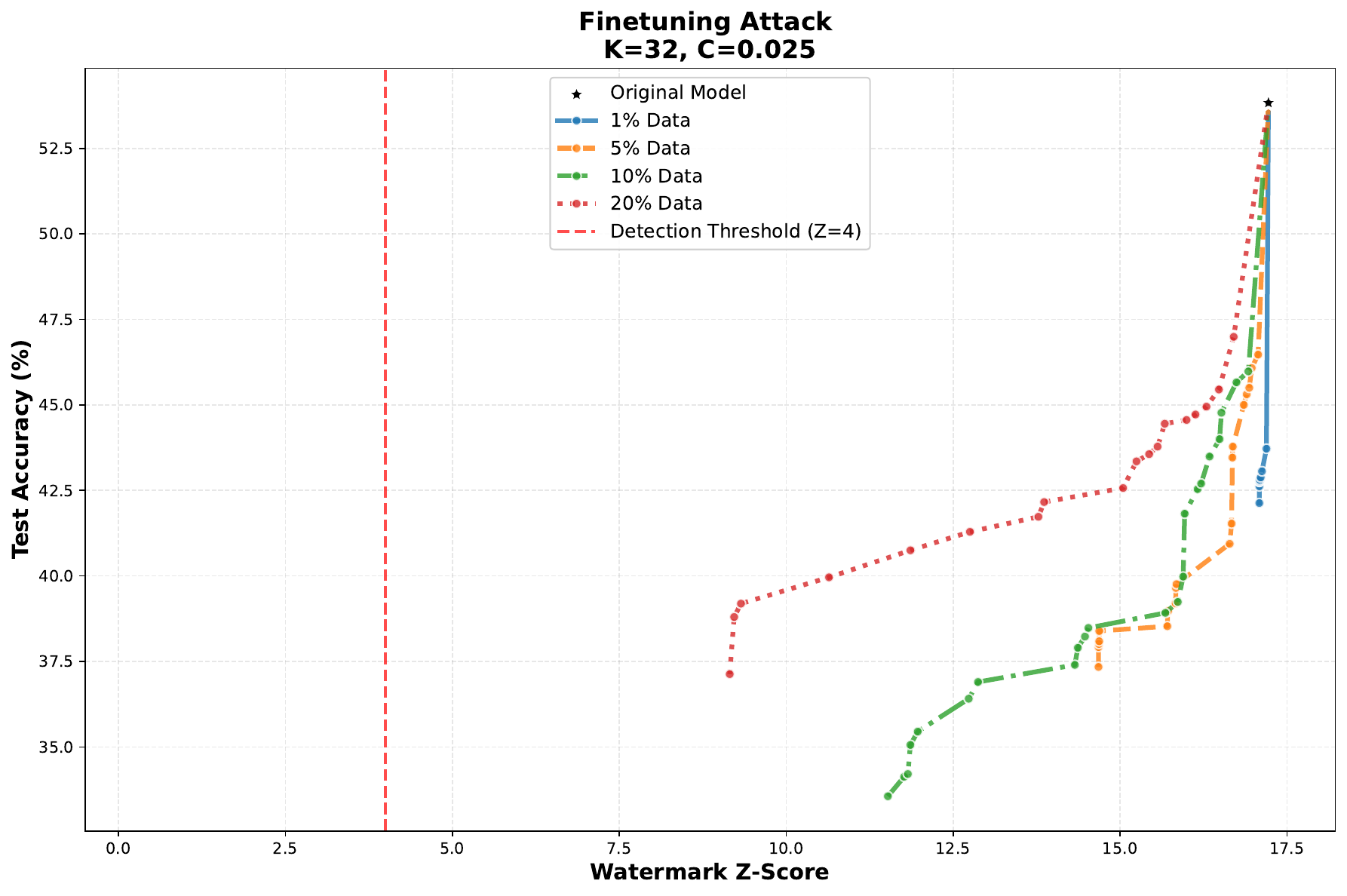}
  \caption{TinyImageNet}
  \label{fig:finetune_tiny}
\end{subfigure}
\caption{Fine-tuning attack: test accuracy versus watermark $z$-score under different fractions of fine-tuning data (5\%, 10\%, 20\%). The dashed red line denotes the detection threshold ($z=4.0$).}
\label{fig:finetune_attack}
\end{figure*}

Fine-tuning gradually reduces the alignment between model parameters and watermark flip vectors, but does not cross the decision threshold within 100 epochs; on the other hand, the model utility is diminished by a very high margin even with 20\% of the available data, as shown in Figure \ref{fig:finetune_attack}.

\subsection{Adaptive Fine-Tuning}
We assume an adaptive attacker who participated during training and estimate the watermark direction by accumulating gradient information from intermediate checkpoints. Let $\tau'$ denote the resulting estimated key, and let $\theta$ denote the current model parameters during fine-tuning. The attacker then fine-tunes the model to reduce its alignment with $\tau'$ by minimizing
\begin{align}
    \mathcal{L}'(\theta)
    = (1-\alpha)\mathcal{L}(\theta)
    + \alpha \,|\cos(\theta,\tau')|
\end{align}

where $\mathcal{L}$ is the model's utility loss, and $\alpha$ controls the tradeoff between the utility loss function and the removal objective. When $\alpha=0$, this reduces to standard fine-tuning. We evaluate $\alpha \in \{0.1,0.3,0.5,0.7\}$. Other hyperparameters (epochs, optimizer, batch size) match fine-tuning. For each epoch, we log accuracy and $z$-score. Pareto frontiers are constructed by retaining non-dominated points in the $(\text{accuracy}, z)$ plane.

Optimizing to minimize the signal of the estimated key improves the effectiveness of the fine-tuning attack. Increasing $\alpha$ drives $z$ down more rapidly, but once $z$ falls below the detection threshold ($z<4$), the model also exhibits very high accuracy degradation.

\begin{figure*}[h]
\centering
\begin{subfigure}[t]{\textwidth}
  \centering
  \includegraphics[width=0.6\linewidth]{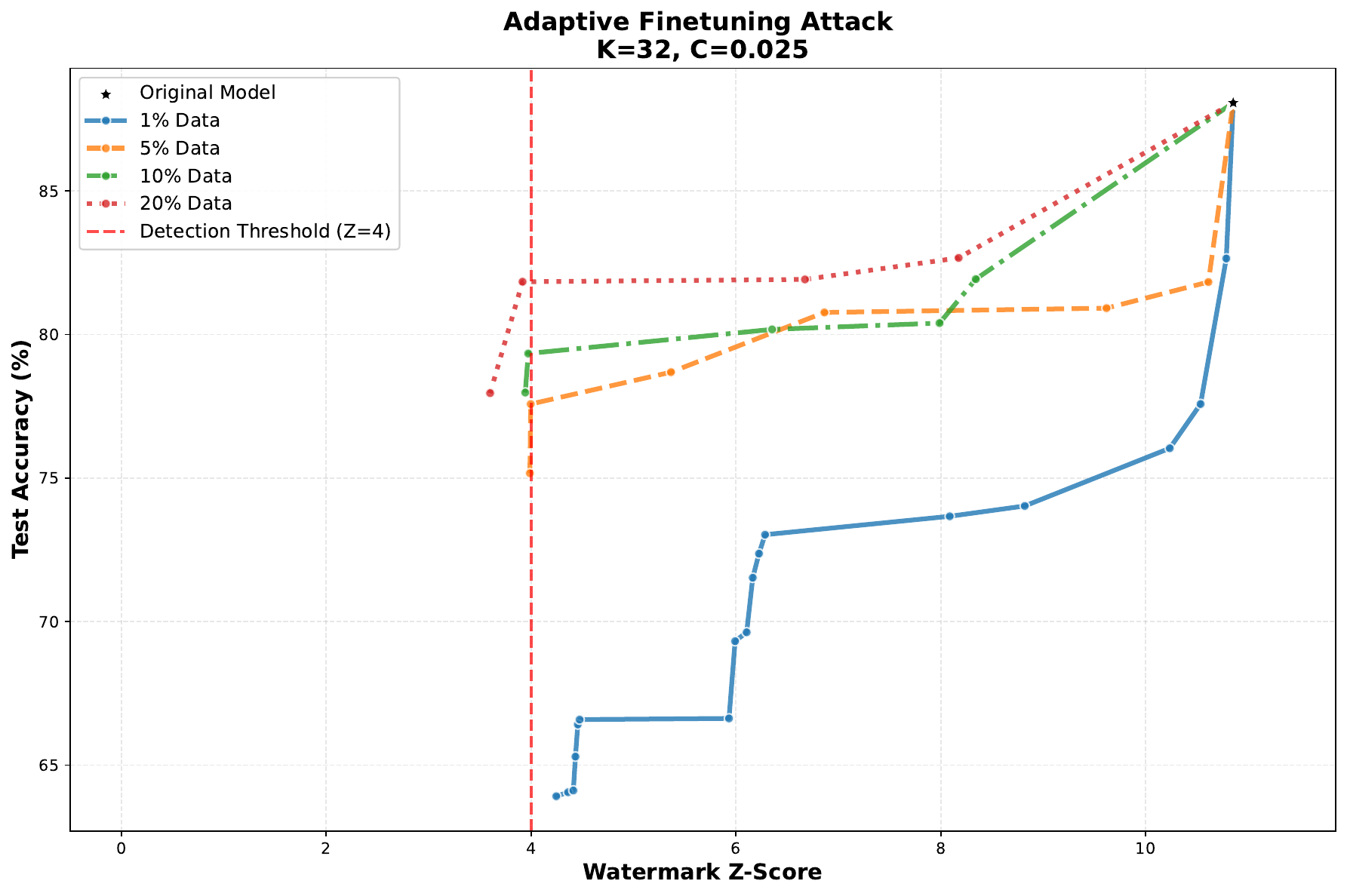}
  \caption{CIFAR-10}
  \label{fig:adaptive_attack_cifar10}
\end{subfigure}
\hfill
\begin{subfigure}[t]{\textwidth}
  \centering
  \includegraphics[width=0.6\linewidth]{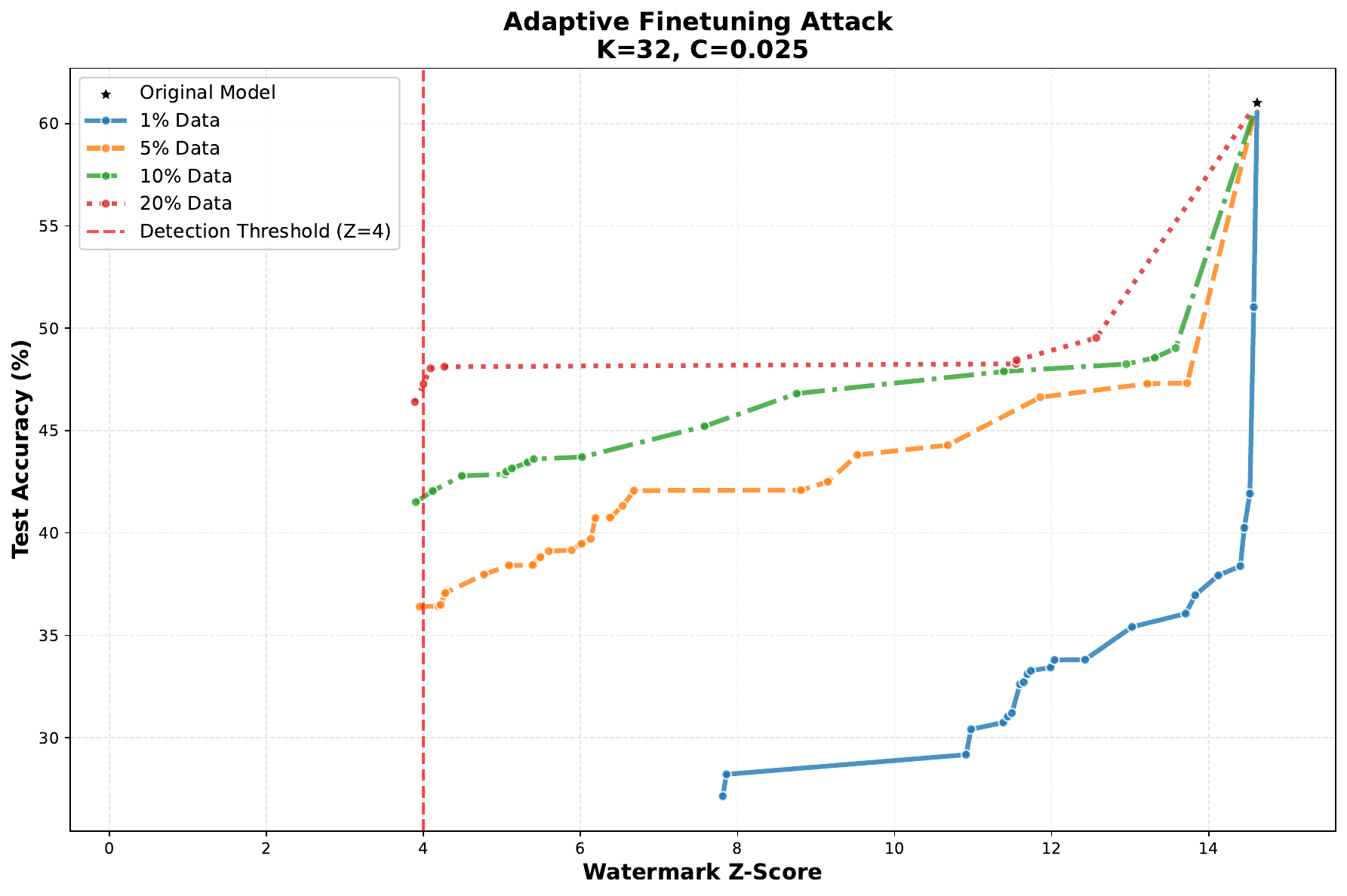}
  \caption{CIFAR-100}
  \label{fig:adaptive_attack_cifar100}
\end{subfigure}
\hfill
\begin{subfigure}[t]{0.6\textwidth}
  \centering
  \includegraphics[width=\linewidth]{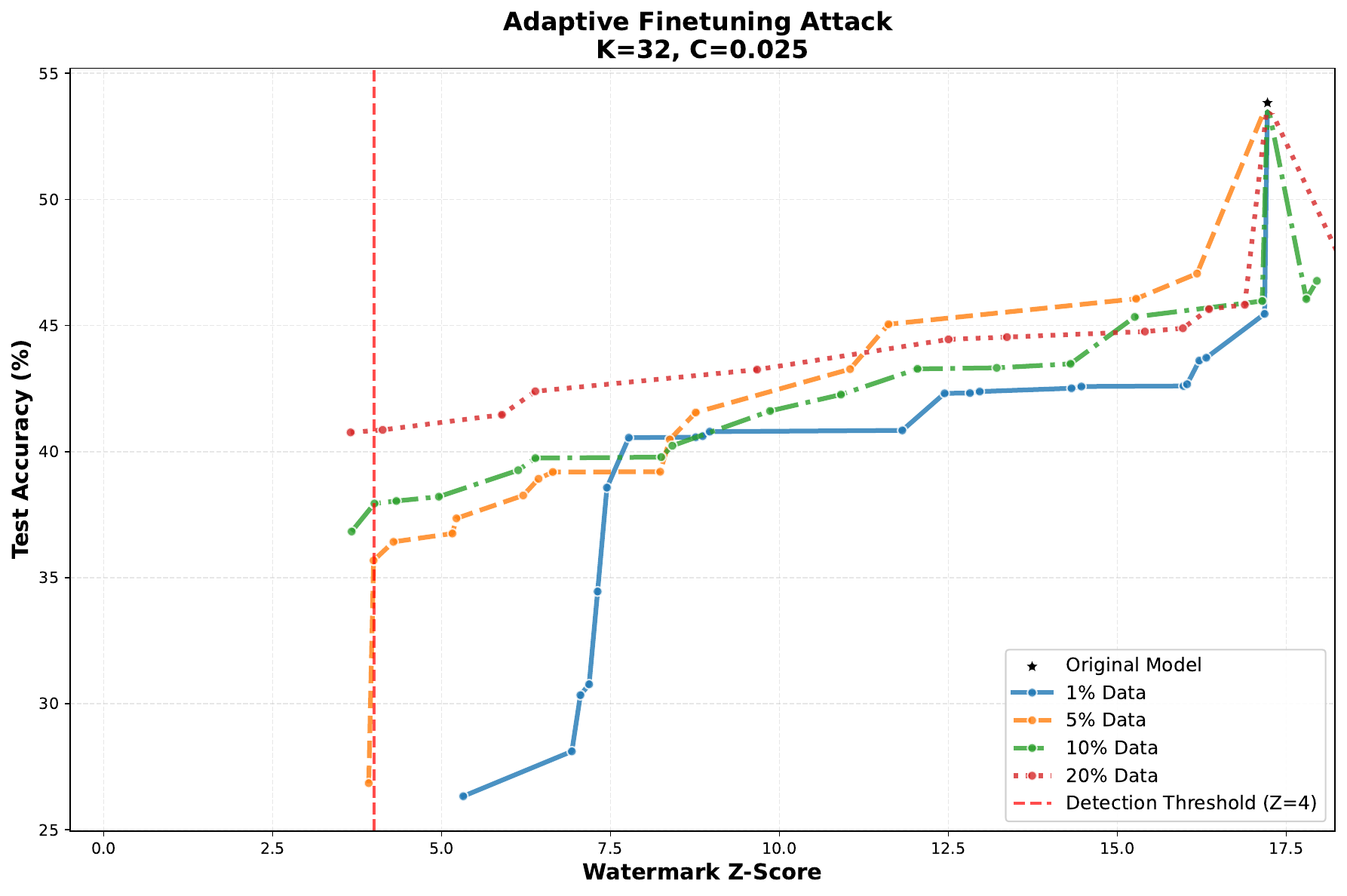}
  \caption{TinyImageNet}
  \label{fig:adaptive_attack_tiny}
\end{subfigure}

\caption{Adaptive fine-tuning attacks with $K=32$ clients and varying watermark strengths ($c \in \{0.075, 0.05, 0.025\}$). Each curve shows the trade-off between test accuracy and watermark detectability ($z$-score) as the adversary fine-tunes the watermarked model with different fractions of training data ($1\%, 5\%, 10\%, 20\%$). The red dashed line marks the detection threshold ($z^* = 4$).}
\label{fig:adaptive_attack}
\end{figure*}

\subsection{Quantization}
Quantization discretizes model weights to a lower numerical precision. Quantization threatens embedded signals because fine-grained correlations in parameter space may be destroyed by discretization.

We apply symmetric weight-only quantization to all Conv/Linear layers, leaving biases and BatchNorm parameters unchanged. Three settings are considered:
\begin{itemize}
    \item \textbf{Static8:} per-tensor symmetric 8-bit quantization,
    \item \textbf{Static4:} per-tensor symmetric 4-bit quantization,
    \item \textbf{Dynamic8:} per-output-channel symmetric 8-bit quantization.
\end{itemize}
For per-tensor quantization, a single scale is computed as the maximum absolute value in the tensor divided by the representable integer range. For per-channel quantization, scales are computed independently for each output channel.

\begin{figure*}[t]
\centering
\begin{subfigure}[t]{0.7\textwidth}
  \centering
  \includegraphics[width=\linewidth]{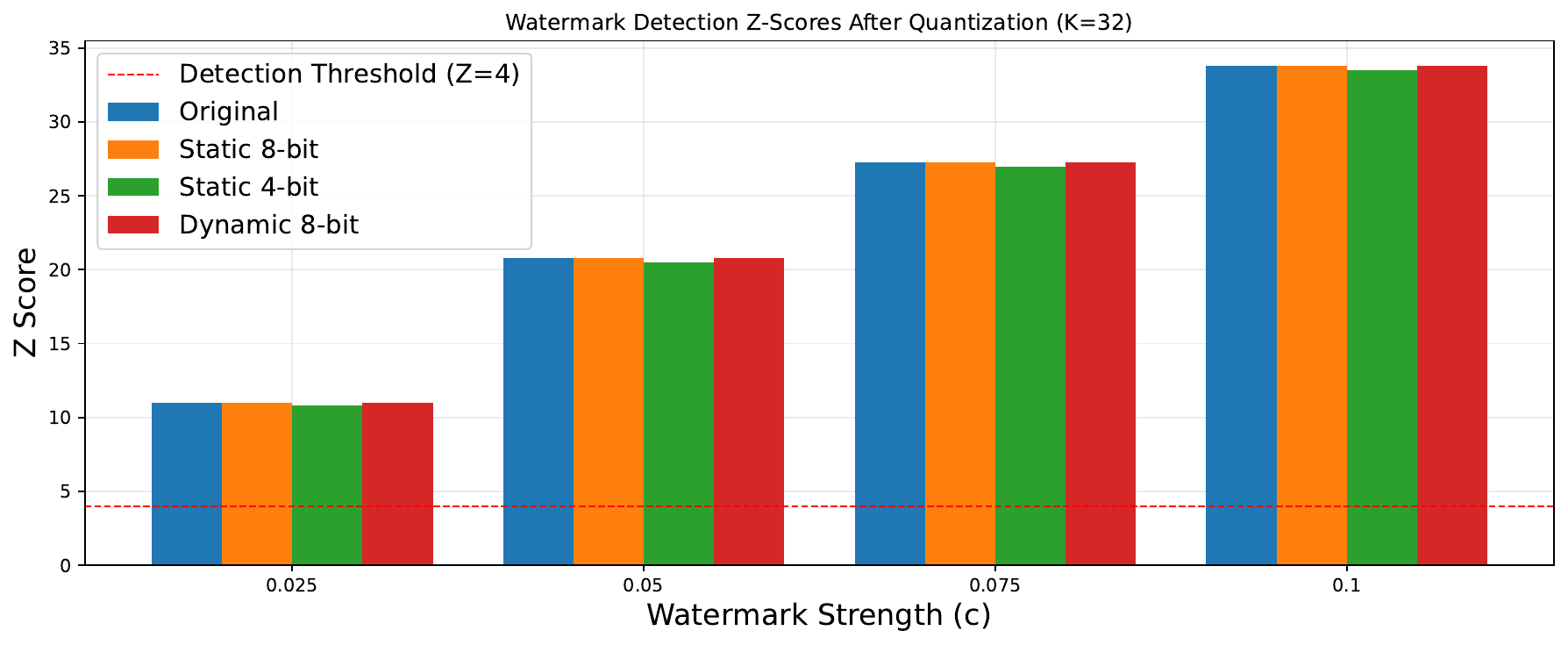}
  \caption{CIFAR-10}
  \label{fig:quant_cifar10}
\end{subfigure}
\hfill
\begin{subfigure}[t]{0.7\textwidth}
  \centering
  \includegraphics[width=\linewidth]{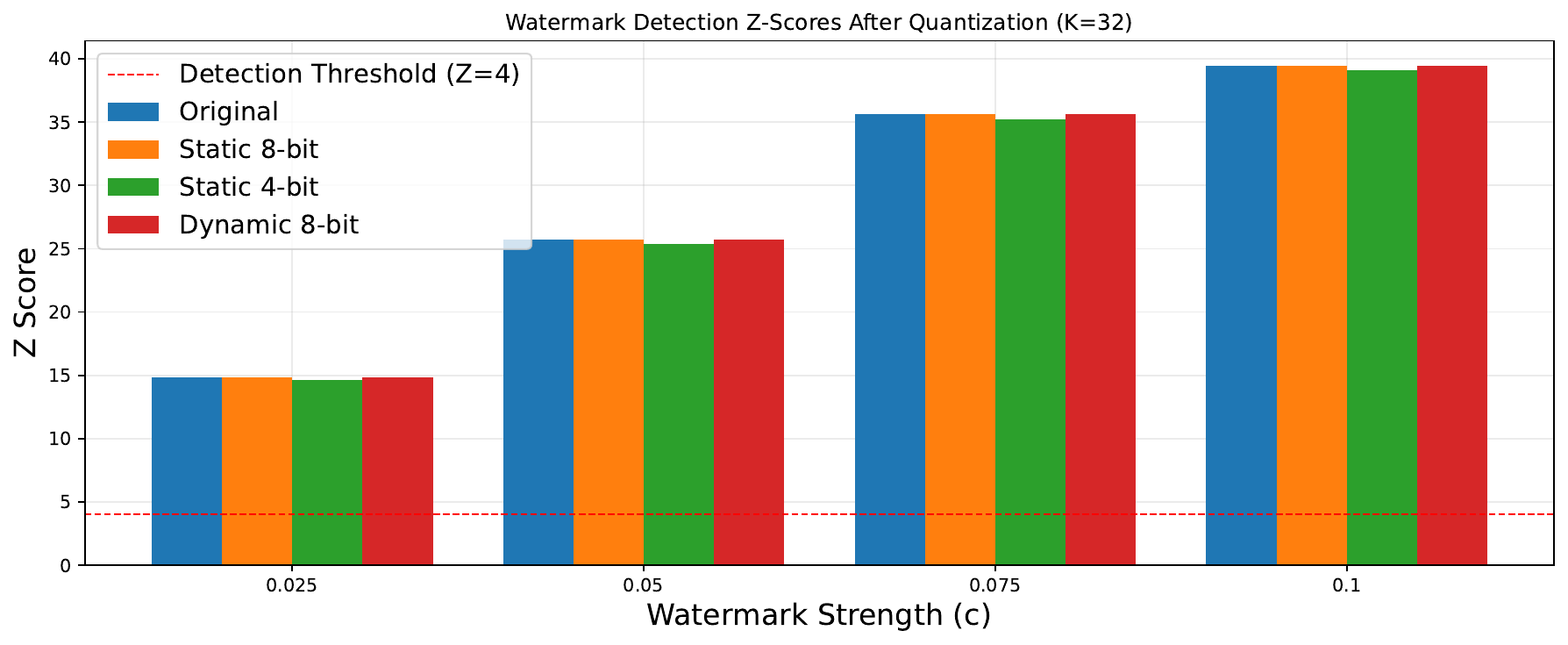}
  \caption{CIFAR-100}
  \label{fig:quant_cifar100}
\end{subfigure}
\hfill
\begin{subfigure}[t]{0.7\textwidth}
  \centering
  \includegraphics[width=\linewidth]{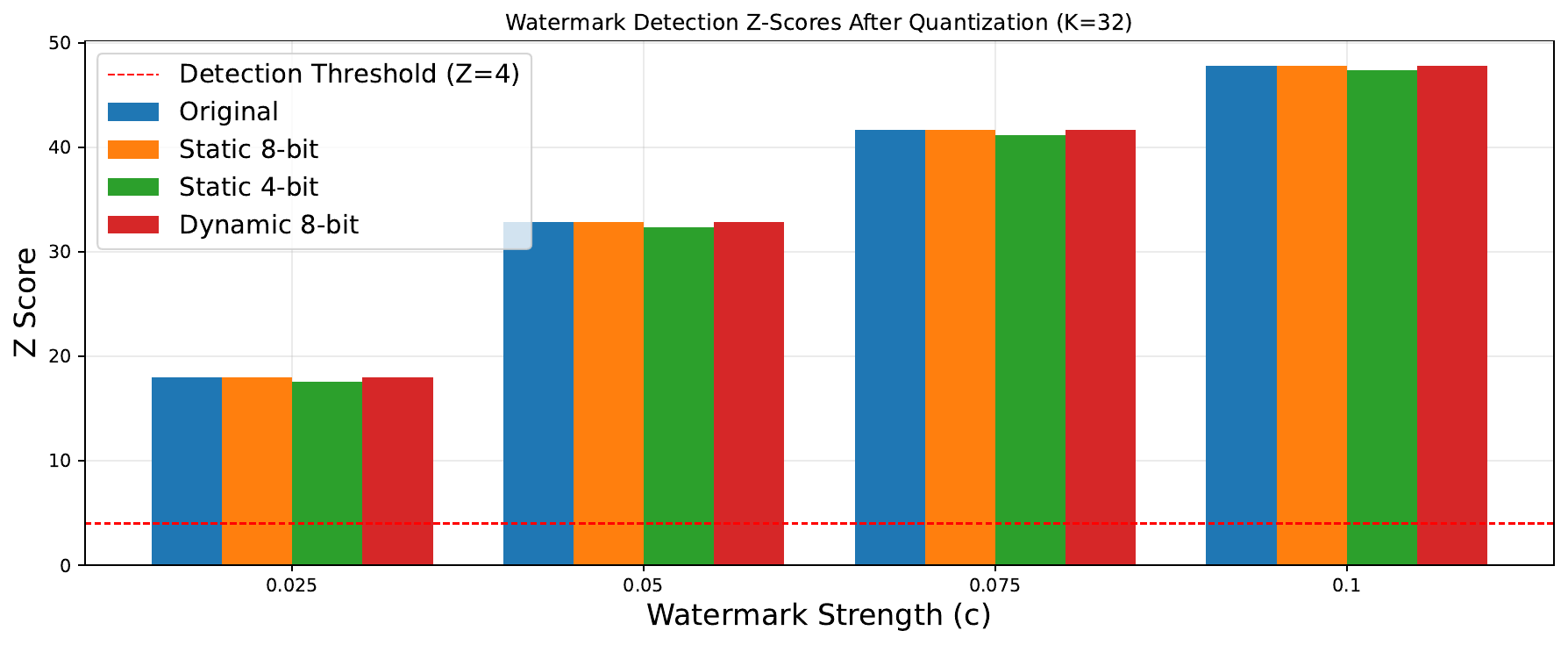}
  \caption{TinyImageNet}
  \label{fig:quant_tiny}
\end{subfigure}

\caption{Watermark detection $z$-scores after quantization. Results are shown for CIFAR-10 (a), CIFAR-100 (b), and TinyImageNet (c) under varying watermark strengths $c$. Each panel compares the original model with three quantization schemes. The dashed red line indicates the detection threshold ($z=4$).}
\label{fig:quantization}
\end{figure*}

Results shown in Figure \ref{fig:quantization} indicate that all quantized models still carry the watermark signal, regardless of the quantization method used. 

\subsection{Pruning}
Pruning reduces model size by eliminating parameters. Both unstructured (sparsity-inducing) and structured (channel-removal) pruning can disrupt embedded watermarks by discarding weights that carry watermark information.
We evaluate two pruning strategies:
\begin{itemize}
    \item \textbf{Magnitude pruning:} global unstructured pruning across all Conv/Linear weights, removing a fraction with the smallest magnitude.
    \item \textbf{Structured pruning:} per-layer pruning of entire output channels, removing a fraction of channels per module.
\end{itemize}
Pruning ratios $\in \{0.3,0.5,0.7,0.9\}$ are tested. Models are evaluated immediately after pruning without retraining. Structured pruning is implemented with $\ell_1$-norm channel selection. 

\begin{figure*}[h]
\centering
\begin{subfigure}[t]{\textwidth}
  \centering
  \includegraphics[width=\linewidth]{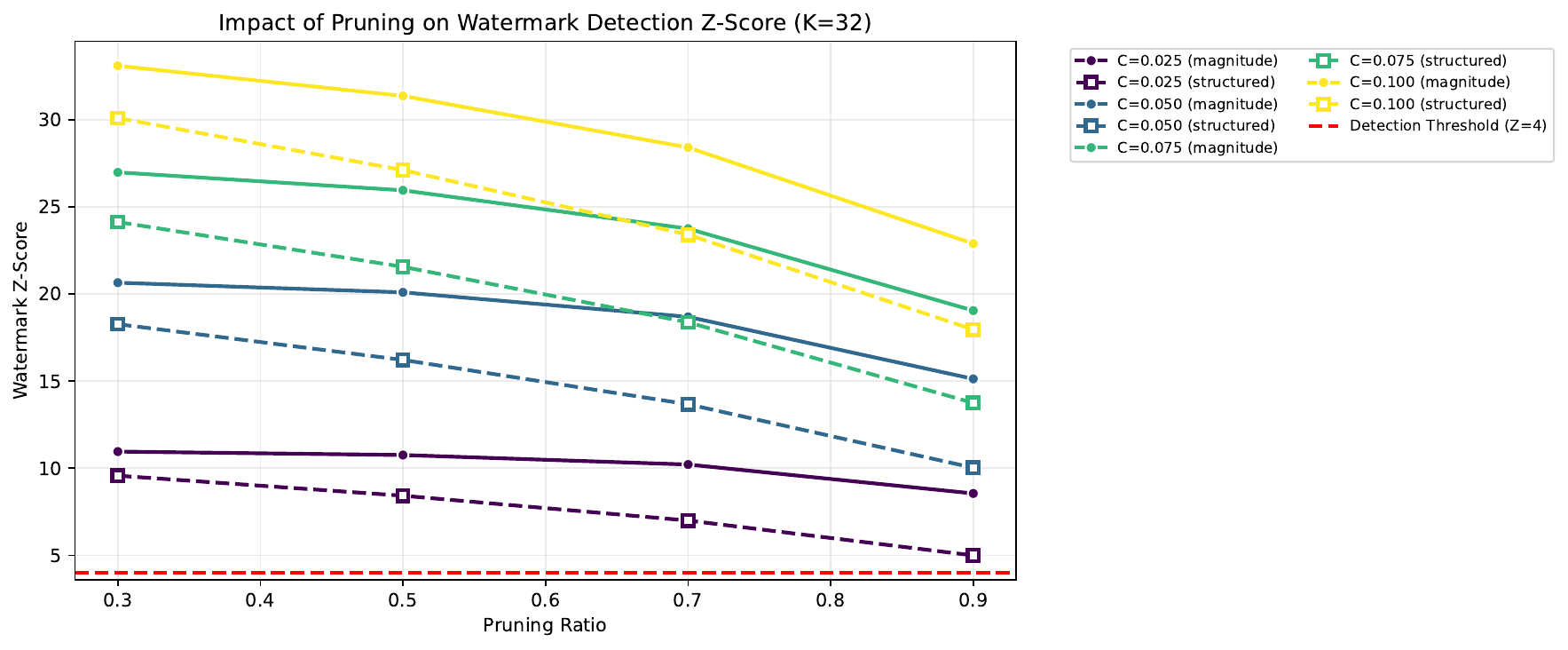}
  \caption{CIFAR-10}
  \label{fig:prune_cifar10}
\end{subfigure}
\hfill
\begin{subfigure}[t]{\textwidth}
  \centering
  \includegraphics[width=\linewidth]{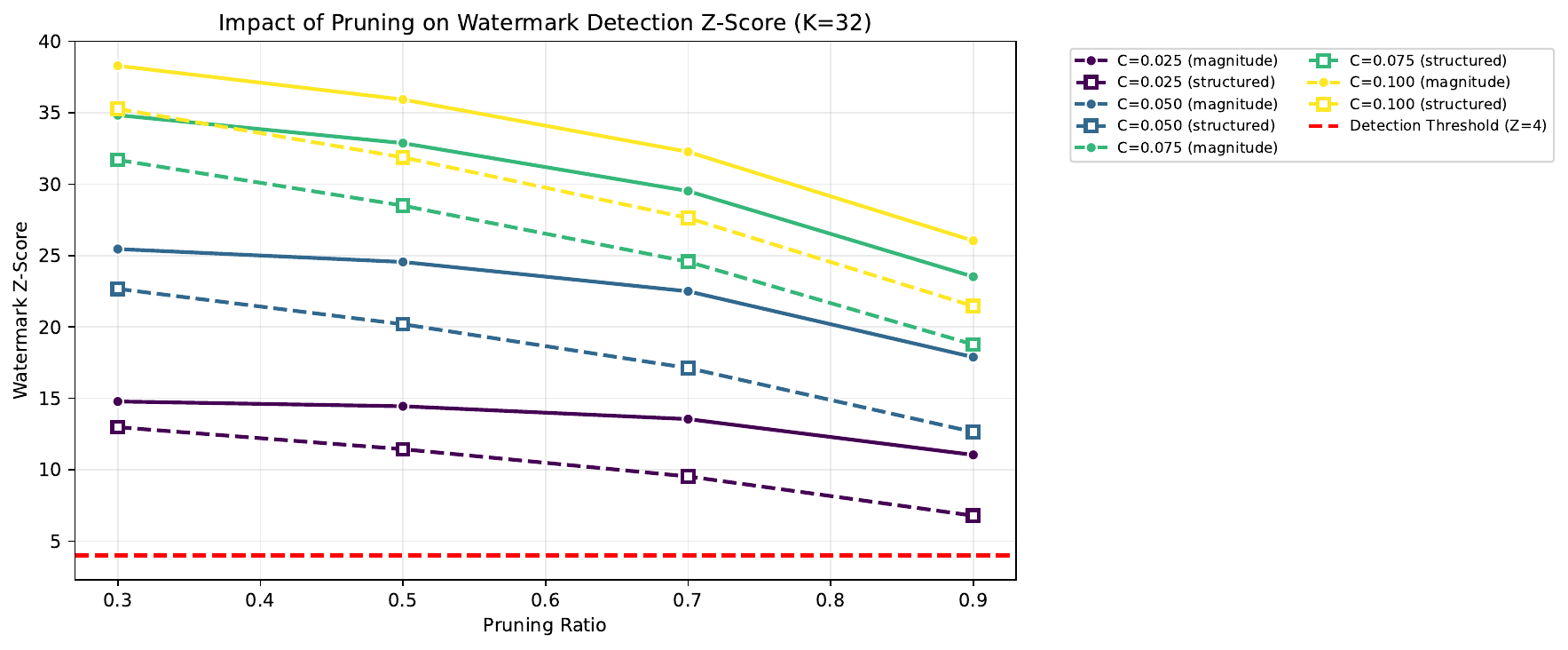}
  \caption{CIFAR-100}
  \label{fig:prune_cifar100}
\end{subfigure}
\hfill
\begin{subfigure}[t]{\textwidth}
  \centering
  \includegraphics[width=\linewidth]{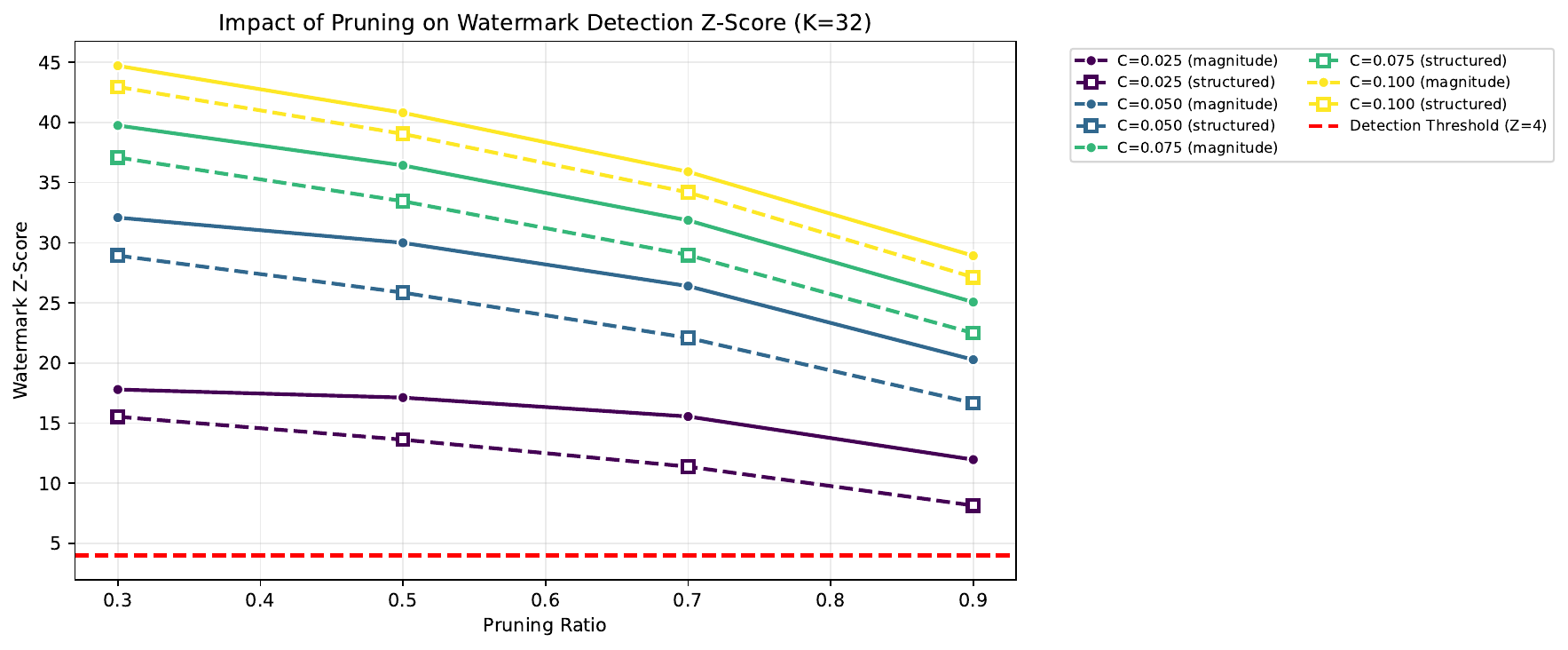}
  \caption{TinyImageNet}
  \label{fig:prune_tiny}
\end{subfigure}

\caption{Impact of pruning on watermark detection $z$-scores across datasets. Results are shown for (a) CIFAR-10, (b) CIFAR-100, and (c) TinyImageNet.}
\label{fig:pruning}
\end{figure*}

Both pruning strategies reduce the watermark signal as the pruning ratio increases, but never cross the decision threshold as shown in Figure \ref{fig:pruning}.

\subsection{Knowledge Distillation}
An attacker can attempt to distill the model by training a student network to imitate the watermarked teacher.

The student is trained using a mixture of ground-truth supervision and distillation from teacher logits. Given temperature $T=3.0$ and weighting $\alpha=0.5$, the loss is
\begin{align}
\mathcal{L} &= \alpha \, \mathrm{KL}\!\Big(\sigma\!\big(\tfrac{z_s}{T}\big) \,||\, \sigma\!\big(\tfrac{z_t}{T}\big)\Big) 
+ (1-\alpha)\, \mathrm{CE}(z_s, y)
\end{align}
where $z_s$ and $z_t$ denote the student and teacher logits.

The student is trained for 100 epochs with Adam (learning rate $1\times10^{-3}$, batch size 128) on subsets $p\% \in \{1,5,10,20\}$ of the exported training data. We report test accuracy, transfer rate, and final watermark $z$-score.

\begin{figure*}[h]
\centering
\begin{subfigure}[t]{0.5\textwidth}
  \centering
  \includegraphics[width=\linewidth]{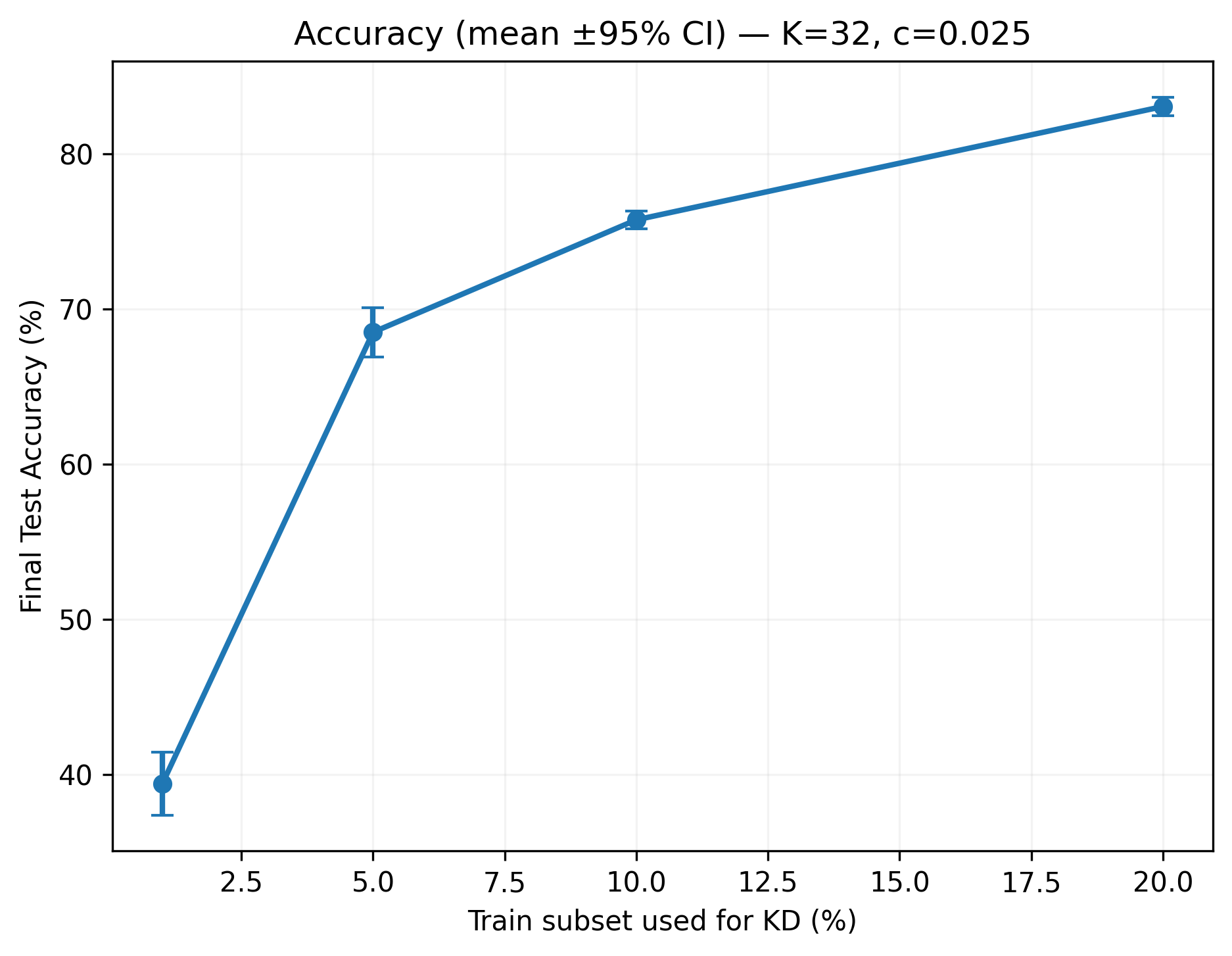}
  \caption{CIFAR-10}
  \label{fig:distill_cifar10}
\end{subfigure}
\hfill
\begin{subfigure}[t]{0.5\textwidth}
  \centering
  \includegraphics[width=\linewidth]{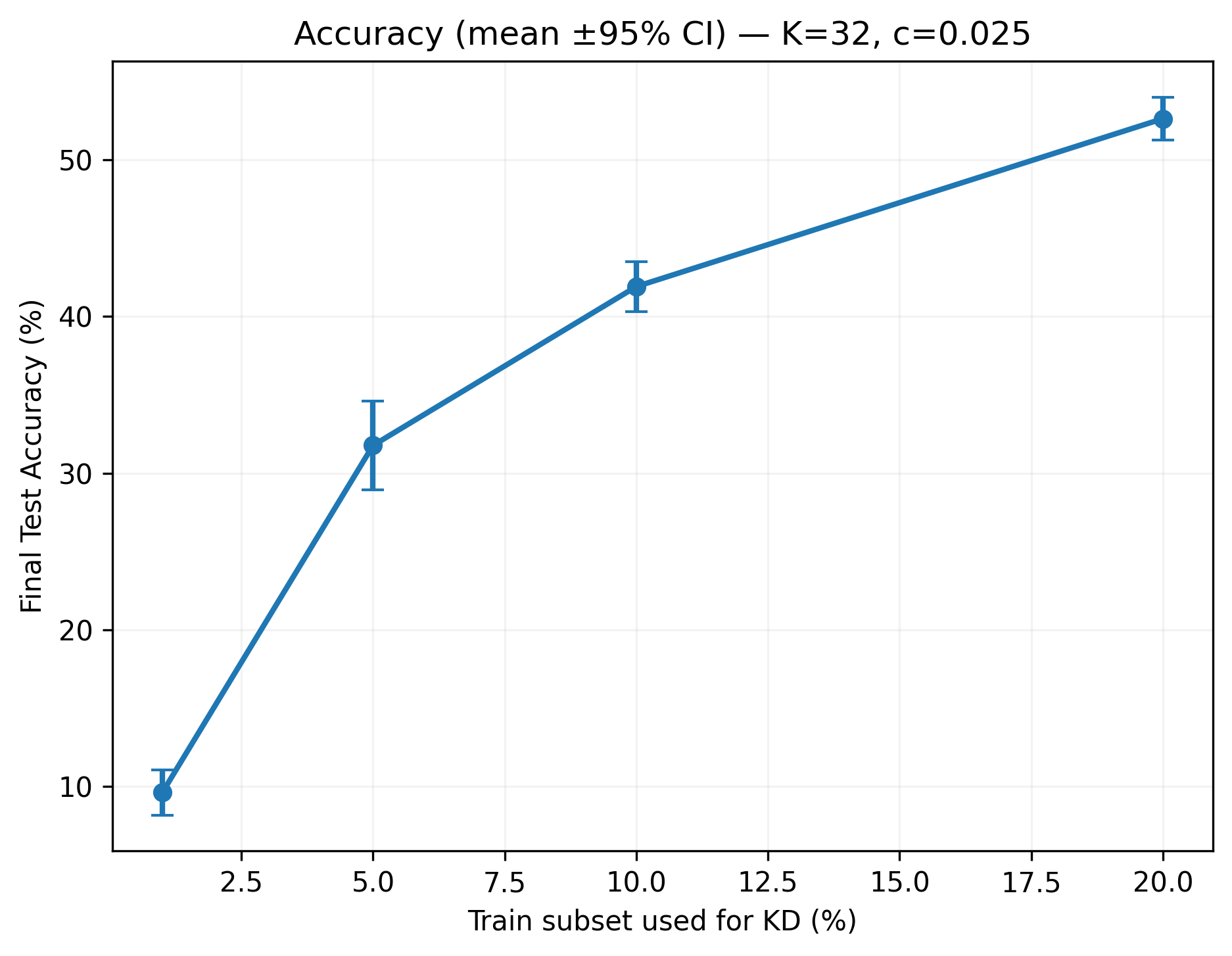}
  \caption{CIFAR-100}
  \label{fig:distill_cifar100}
\end{subfigure}
\hfill
\begin{subfigure}[t]{0.5\textwidth}
  \centering
  \includegraphics[width=\linewidth]{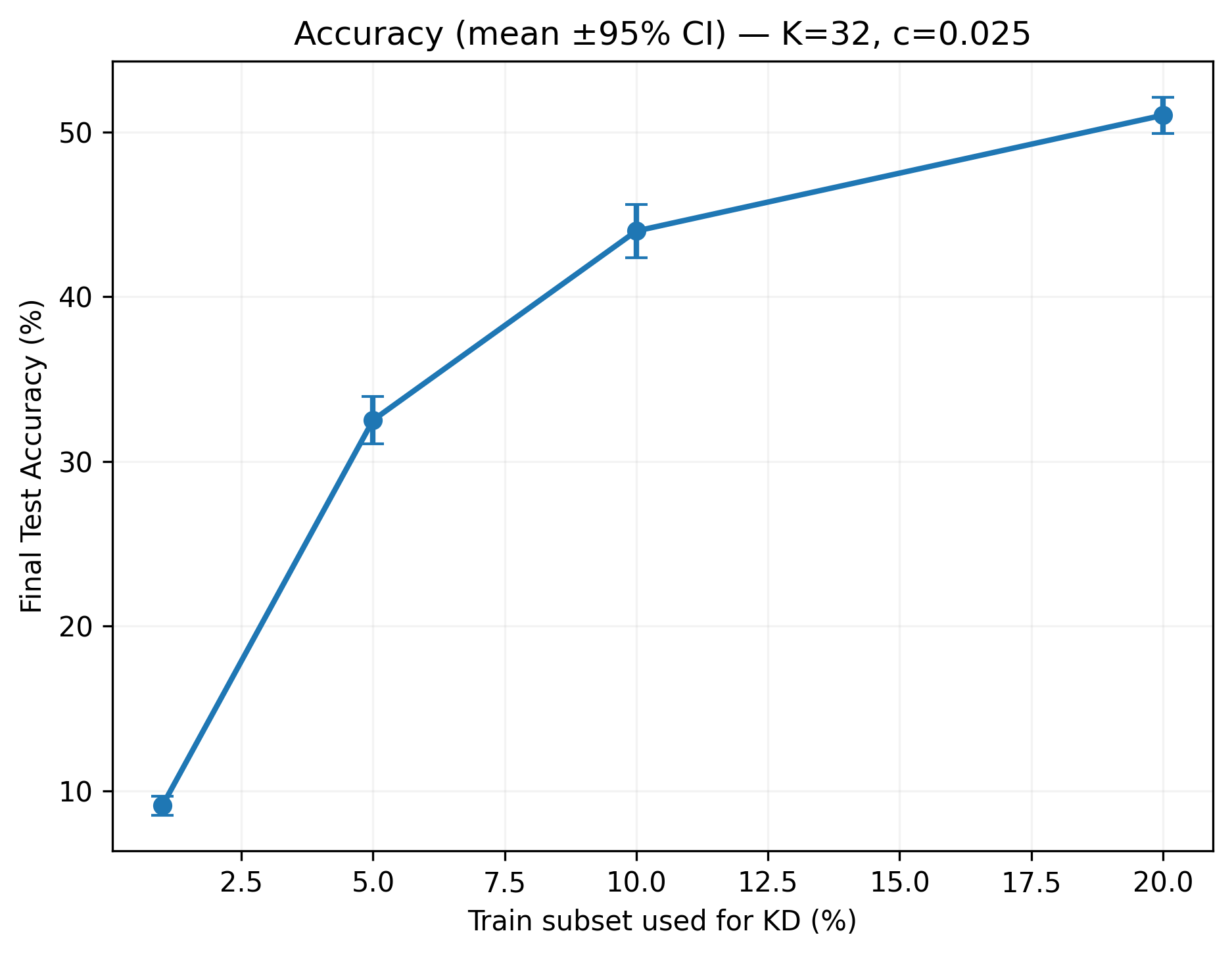}
  \caption{TinyImageNet}
  \label{fig:distill_tiny}
\end{subfigure}

\caption{Distillation attack with $K=32$ clients and watermark strength $c=0.025$. Final test accuracy is shown as the fraction of training data used for knowledge distillation increases. Results are reported for (a) CIFAR-10, (b) CIFAR-100, and (c) TinyImageNet. Error bars denote 95\% confidence intervals across seeds.}
\label{fig:distillation}
\end{figure*}

Distillation can produce student models that approach the teacher’s task accuracy while noticeably attenuating the watermark signal, since the functional behavior is transferred without preserving parameter-level correlations. However, we observe that distillation doesn't produce a high utility model when the available data fraction is below 20\%, illustrated in Figure \ref{fig:distillation}, meaning that a substantial portion of the training data is required for the attack to succeed. Moreover, training a new student network is computationally expensive, making this approach significantly more costly than lightweight post-processing attacks such as pruning or quantization.

\subsection{Pareto graphs}
We extend the accuracy--z-score Pareto analysis to CIFAR-10 and Tiny ImageNet. Across both datasets, the watermark exhibits the same robustness patterns observed in the main CIFAR-100 results: post-training attacks must induce a clear degradation in task accuracy before reducing the watermark statistic below the detection threshold ($z \geq 4$). In particular, quantization has negligible impact on detectability, while fine-tuning and pruning trade accuracy for only gradual reductions in z-score. These results confirm that the observed robustness behavior is consistent across datasets of varying difficulty and scale.

\begin{figure*}[h]
  \centering
  \includegraphics[width=\textwidth]{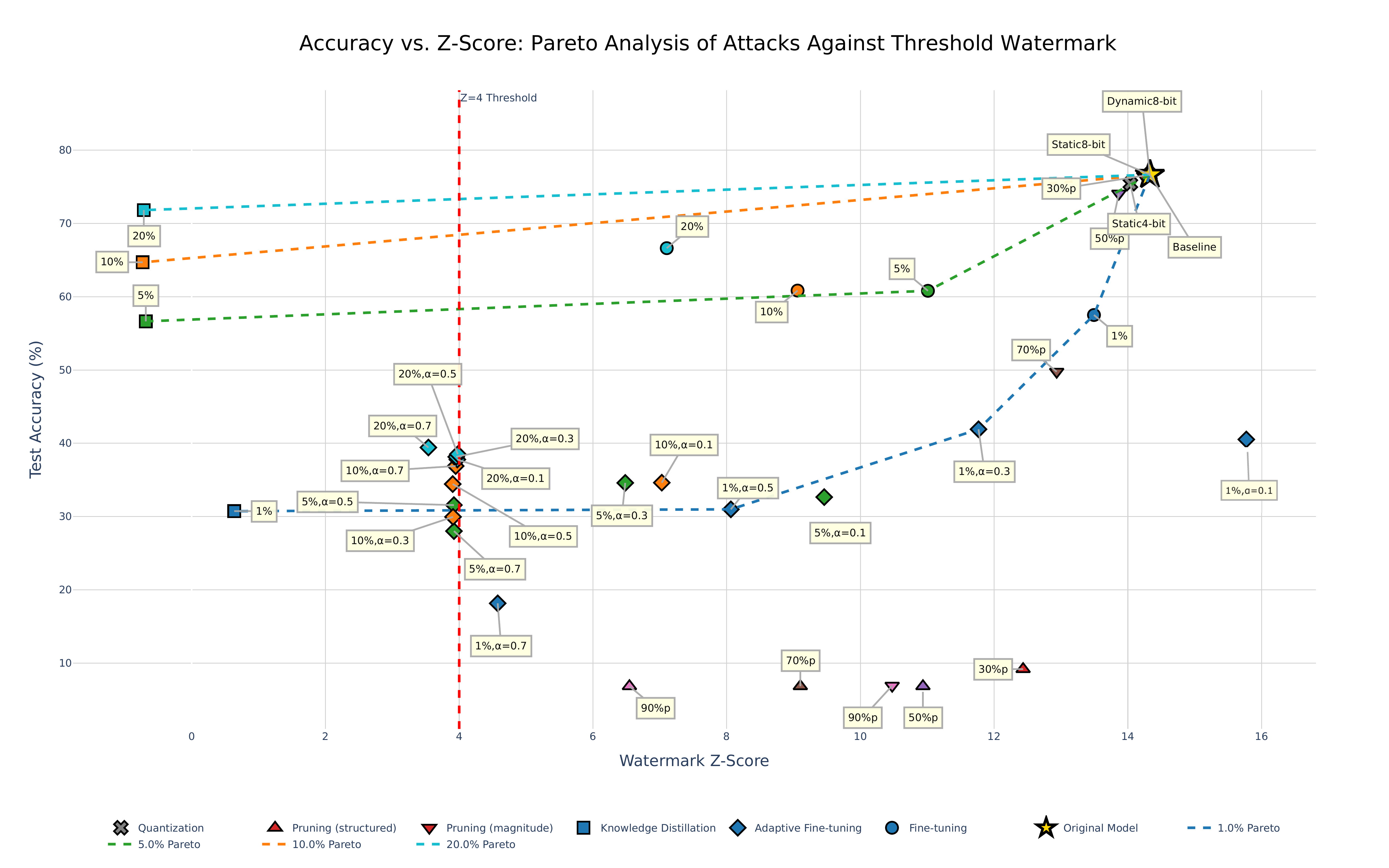}
  \caption{Robustness analysis on Tinyimagenet with $K=32$ and $c=0.025$. 
  We report the trade-off between task accuracy and watermark $z$-score under five attack types: (i) adaptive fine-tuning, (ii) plain fine-tuning, (iii) knowledge distillation, (iv) pruning (magnitude and structured), and (v) quantization. The original model is shown as a star. Dashed curves denote Pareto frontiers for 1\%, 5\%, 10\%, and 20\% of the training data, while the red dashed line marks the detection threshold ($z=4$).}
  \label{fig:robustness_tiny}
\end{figure*}

\begin{figure*}[h]
  \centering
  \includegraphics[width=\textwidth]{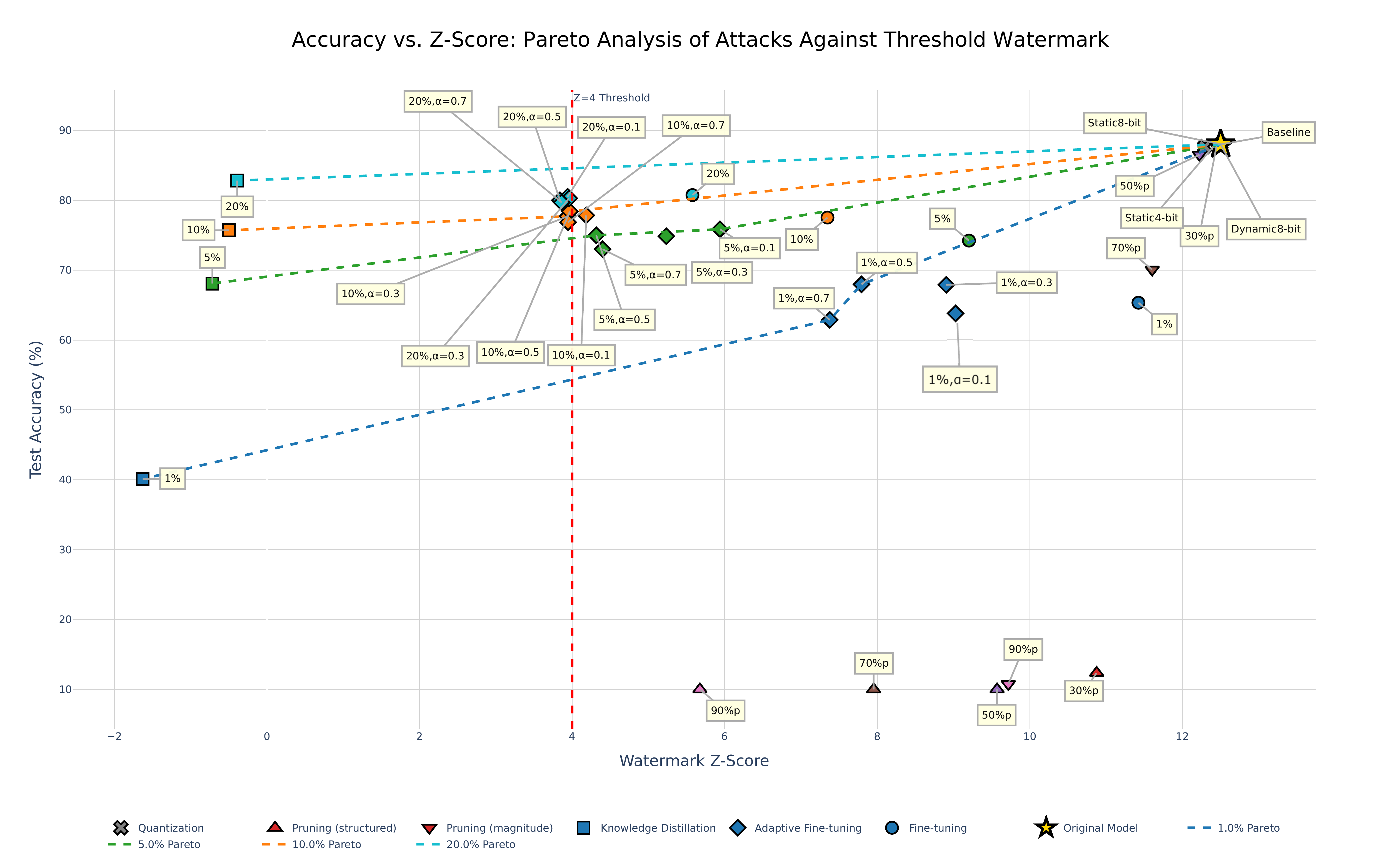}
  \caption{Robustness analysis on CIFAR-10 with $K=32$ and $c=0.025$. 
  We report the trade-off between task accuracy and watermark $z$-score under five attack types: (i) adaptive fine-tuning, (ii) plain fine-tuning, (iii) knowledge distillation, (iv) pruning (magnitude and structured), and (v) quantization. The original model is shown as a star. Dashed curves denote Pareto frontiers for 1\%, 5\%, 10\%, and 20\% of the training data, while the red dashed line marks the detection threshold ($z=4$).}
  \label{fig:robustness_cifar10}
\end{figure*}


\end{document}